\documentclass[11pt,a4paper]{article}
\usepackage[hyperref]{naaclhlt2019}
\usepackage{times}
\usepackage{latexsym}

\usepackage{url}

\usepackage{graphicx}   
\usepackage{subfigure}
\usepackage{multirow}
\usepackage{booktabs}
\usepackage{diagbox}
\usepackage{color}
\usepackage{bbm}
\usepackage{multicol}
\usepackage{blindtext}
\usepackage{amsmath,amsfonts,amssymb}
\usepackage{bm}
\usepackage{mathtools}
\usepackage{wrapfig}
\usepackage{wrapfig,lipsum,booktabs}
\usepackage{url}
\usepackage{algorithm}
\usepackage{algpseudocode}

\DeclareMathOperator*{\softmax}{softmax}
\DeclareMathOperator*{\sigmoid}{sigmoid}
\DeclareMathOperator*{\relu}{relu}
\DeclareMathOperator*{\identity}{identity}
\DeclareMathOperator*{\elu}{elu}
\DeclareMathOperator*{\dpattn}{sdpAttn}

\DeclareMathOperator*{\tsa}{TSA}
\DeclareMathOperator*{\dropout}{Dropout}

\graphicspath{{./figures/}}

\aclfinalcopy 

\title{Tensorized Self-Attention: \\Efficiently Modeling Pairwise and Global Dependencies Together}

\author{
Tao Shen$^1$, 
Tianyi Zhou$^2$, 
Guodong Long$^1$, 
Jing Jiang$^1$, 
Chengqi Zhang$^1$
\\ 
$^1$ Centre for Artificial Intelligence, School of Software, University of Technology Sydney \\
$^2$ Paul G. Allen School of Computer Science \& Engineering, University of Washington\\
tao.shen@student.uts.edu.au, tianyizh@uw.edu, guodong.long@uts.edu.au, \\
jing.jiang@uts.edu.au, chengqi.zhang@uts.edu.au
}

\date{}

\begin{document}
\maketitle
\begin{abstract}
  Neural networks equipped with self-attention have parallelizable computation, light-weight structure, and the ability to capture both long-range and local dependencies. Further, their expressive power and performance can be boosted by using a vector to measure pairwise dependency, but this requires to expand the alignment matrix to a tensor, which results in memory and computation bottlenecks. In this paper, we propose a novel attention mechanism called ``Multi-mask Tensorized Self-Attention'' (MTSA), which is as fast and as memory-efficient as a CNN, but significantly outperforms previous CNN-/RNN-/attention-based models. MTSA 1) captures both pairwise (token2token) and global (source2token) dependencies by a novel compatibility function composed of dot-product and additive attentions, 2) uses a tensor to represent the feature-wise alignment scores for better expressive power but only requires parallelizable matrix multiplications, and 3) combines multi-head with multi-dimensional attentions, and applies a distinct positional mask to each head (subspace), so the memory and computation can be distributed to multiple heads, each with sequential information encoded independently. The experiments show that a CNN/RNN-free model based on MTSA achieves state-of-the-art or competitive performance on nine NLP benchmarks with compelling memory- and time-efficiency. 
\end{abstract}

\section{Introduction}

Recurrent neural network (RNN) and convolutional neural network (CNN) have been broadly used as context fusion modules for natural language processing (NLP) tasks.
Recently, RNN/CNN in conjunction with an attention mechanism has been proven to be effective for contextual feature modeling in a wide range of NLP tasks, including sentiment classification \cite{lizheng2018hierarchical}, machine translation \cite{bahdanau2015neural}, reading comprehension \cite{seo2017bidirectional,yu2018qanet}, etc. 
More recently, self-attention mechanisms have been developed for context fusion and syntactic dependency modeling with the advantage of fewer parameters, more parallelizable computation, and better empirical performance \cite{hu2017reinforced,vaswani2017attention,shen2017disan}. In addition, neural networks based solely on self-attention mechanisms have achieved state-of-the-art quality on many NLP tasks, e.g., machine translation \cite{vaswani2017attention}, sentence embedding \cite{shen2017disan} and semantic role labeling \cite{tan2017deep}. 

Self-attention mechanisms can be categorized into two classes according to the type of dependency each aims to model. The first category is \emph{token2token} self-attention \cite{hu2017reinforced,vaswani2017attention,shen2017disan} that captures syntactic dependency between every two tokens in a sequence. An efficient dot-product compatibility function is usually deployed to measure this pairwise dependency \cite{vaswani2017attention}. In contrast, additive compatibility function captures the dependency by multi-layer perceptron (MLP), and can usually achieve better performance \cite{britz2017massive}. 
Its expressive power can be further improved if expanded to multiple dimensions \cite{shen2017disan}. This multi-dim self-attention empirically surpasses dot-product one, but suffers from expensive computation and memory, which grow linearly with the number of features and quadratically with the sequence length. Hence, it is not scalable to long sequences in practice. 

\begin{figure*}[t]
	\centering
	\subfigure[\small]{
		\label{fig:line_chart_memory}
		\includegraphics[width=0.27\textwidth]{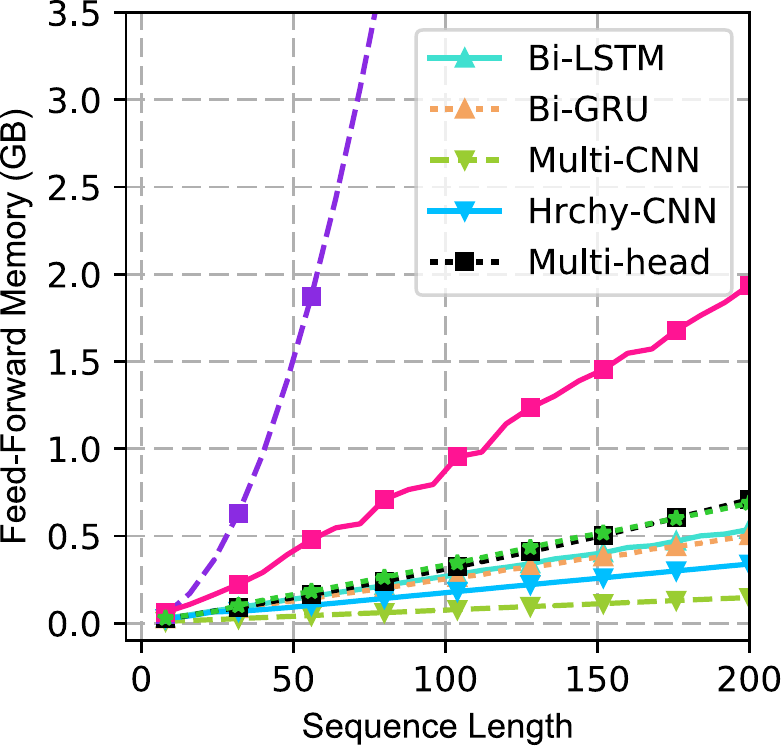}}  
	\hspace{5mm}
	\subfigure[\small]{
		\label{fig:line_chart_time}
		\includegraphics[width=0.2745\textwidth]{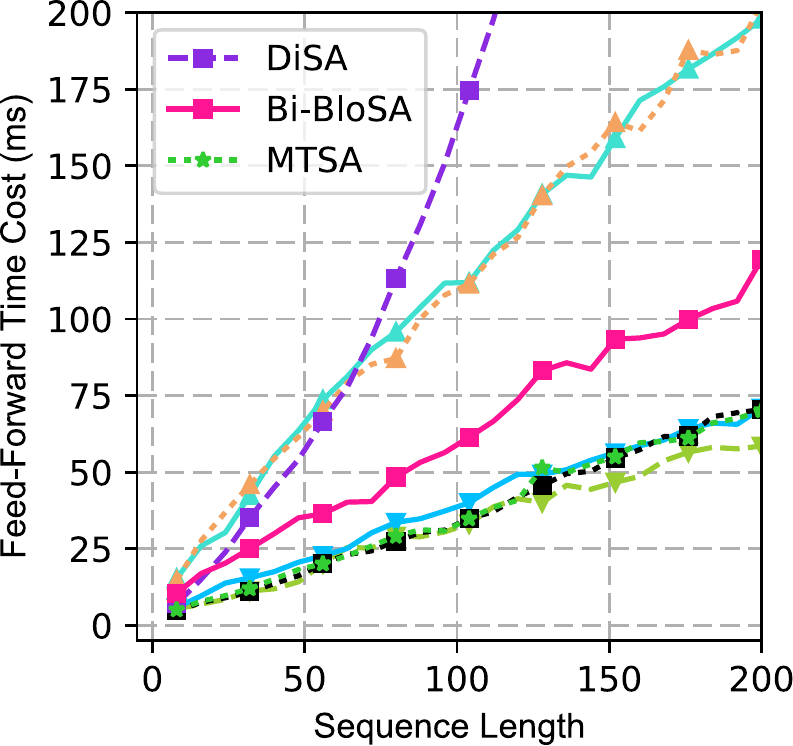}} 
	\hspace{5mm}
	\subfigure[\small]{
		\label{fig:scatter_chart_snli}
		\includegraphics[width=0.2475\textwidth]{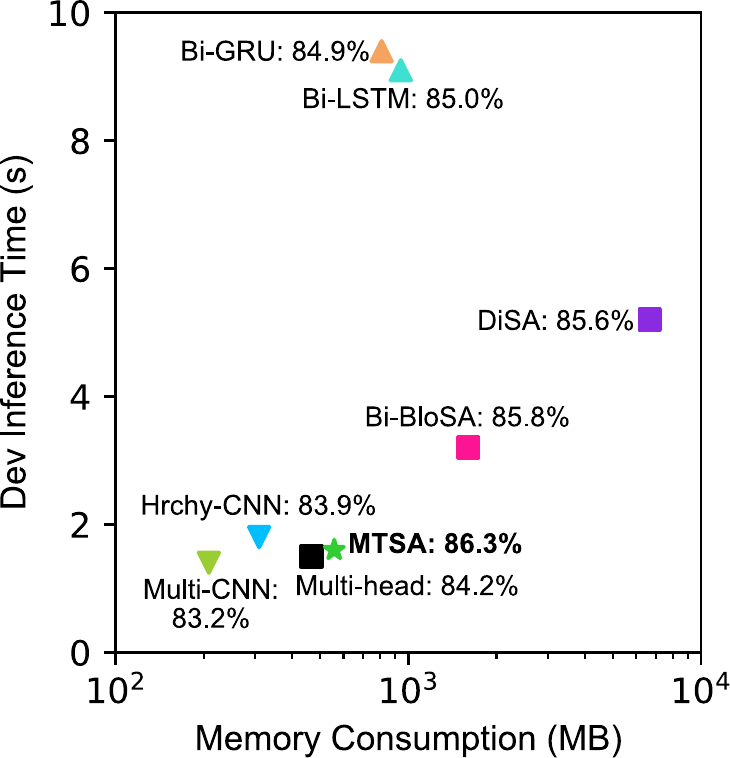}} 
	\caption{(a) Memory consumption and (b) time cost \textit{vs.} sequence length on synthetic data; (c) memory load ($x$-axis), inference time on dev set ($y$-axis) and test accuracy on the SNLI dataset.}
	\label{fig:intro_collection}
\end{figure*}

The second category is \emph{source2token} self-attention \cite{liu2016learning,lin2017structured,shen2017disan} aiming to capture global dependency, i.e., the importance of each token to the entire sequence for a specific task. Its time and space complexities grow linearly, rather than quadratically, with the sequence length. Hence, it is empirically efficient in terms of memory and computation even if expanded to multiple dimensions, i.e., using a vector of feature-wise scores instead of a scalar for the global dependency. But, it is hard to reach state-of-the-art performance on NLP tasks due to the lack of pairwise and local dependencies.

In this paper, we propose a novel attention mechanism called multi-mask tensorized self-attention (MTSA)\footnote{More details about training setups, related work, discussion, and visualization are provided in the Appendix.}, for context fusion. In MTSA,  
1) the pairwise dependency is captured by an efficient dot-product based token2token self-attention, while the global dependency is modeled by a feature-wise multi-dim source2token self-attention, so they can work jointly to encode rich contextual features; 
2) self-attention alignment scores are tensorized for more expressive power in that each pair of tokens has one score for each feature, but no tensor computation is required other than simple and efficient matrix multiplications when implemented;
3) the tensors above are computed in multiple subspaces (i.e., in a multi-head fashion) rather than in the original input space, so the required memory and computation can be distributed to multiple subspaces; and 
4) a distinct positional mask is applied to each head in order to encode rich structural information such as the sequential order and relative position of tokens.

In the experiments, we build CNN/RNN-free neural networks based on MTSA for sentence embedding and sequence tagging tasks, including natural language inference, semantic role labeling, sentiment analysis, question-type classification, machine translation, etc. 
The results demonstrate that  MTSA achieves state-of-the-art or competitive performance on nine benchmark datasets. 
To summarize the comparison of MTSA with recently popular models, we show the memory consumption and time cost \textit{vs.} sequence length respectively in Figure \ref{fig:line_chart_memory} and \ref{fig:line_chart_time} on synthetic data (batch size of 64 and feature channels of 300). On the SNLI~\cite{bowman2015snli}, a public dataset for language inference, as shown in Figure \ref{fig:scatter_chart_snli}, MTSA achieves the best result but is as fast and as memory-efficient as the CNNs (all baselines and the benchmark are detailed in Section \ref{sec:experiments}). 

\textbf{Notations:} 1) lowercase denotes a vector; 2) bold lowercase denotes a sequence of vectors (stored as a matrix); and 3) uppercase denotes a matrix or tensor. 

\section{Background}
\subsection{Attention Mechanism}

Given an input sequence of token embeddings or memory slots $\bm{x} = [x_1, \dots,x_n] \in \mathbb{R}^{d_e \times n}$, and a vector representation of a query $q\in \mathbb{R}^{d_q}$, attention mechanism \cite{bahdanau2015neural,luong2015effective} computes an alignment score between each token $x_i$ and $q$ by a compatibility function $f(x_i, q)$, which aims to measure the dependency/relevance between $x_i$ and $q$, or the attention of $q$ to $x_i$, w.r.t. a given task. The scores are transformed into probabilities through a $\softmax$ function. These probabilities are then used as weights to sum all the tokens and generate a contextual embedding for $q$, i.e.,  
\begin{align} 
	\notag &p(z|\bm{x}, q) = \softmax(a),~~a= [f(x_i, q)]_{i=1}^{n}, \\
	&s = \sum_{i=1}^{n} p(z=i|\bm{x}, q) \cdot x_i = \mathbb{E}_{i \sim p(z|\bm{x}, q)}[x_i], 
\end{align}
where $a\in\mathbb{R}^{n}$ denotes the vector of $n$ alignment scores, $p(z|\bm{x}, q)$ is the categorical distribution for attention probabilities, which is derived from applying $\softmax$ function to $a$. And, $s\in \mathbb{R}^{d_e}$ is the output vector for the query $q$.

There are two major types of compatibility functions, leading to the two most frequently used attention mechanisms. The first one is dot-product or multiplicative compatibility function (Eq.(\ref{eq:comp_functions_1})), which composes \textbf{dot-product attention} mechanism \cite{luong2015effective} using cosine similarity to model the dependencies. The other one is additive or multi-layer perceptron (MLP) compatibility function (Eq.(\ref{eq:comp_functions_2})) that results in \textbf{additive attention} mechanism \cite{bahdanau2015neural} using MLP to model the dependencies. 
\begin{align}
	&f(x_i, q) = \langle W^{(d1)} x_i, W^{(d2)} q \rangle, \label{eq:comp_functions_1}\\
	&f(x_i, q) = w^{T} \sigma_a (W^{(a)}[x_i; q] + b^{(a)})+b, \label{eq:comp_functions_2}
\end{align}
where $W^{(d1)} \in \mathbb{R}^{d_i\times d_e}, W^{(d2)} \in \mathbb{R}^{d_i\times d_q}, W^{(a)} \in \mathbb{R}^{d_a\times (d_e\!+\!d_q)}, w \in \mathbb{R}^{d_a}$ are learnable parameters, $\langle \cdot ,\cdot \rangle$ denotes inner-product. Empirically, networks with additive attention usually outperform those with dot-product attention, but require more computation time and memory \cite{britz2017massive}. 

\textbf{Multi-dim attention} mechanism \cite{shen2017disan} expands the alignment score in previous attention mechanisms to a vector for feature-wise scores, each computed on a feature dimension. It has greater capacity to model complex dependencies, and can handle context variation and polysemy problems harassing many NLP tasks. \textbf{In particular, it replaces vector $w^T\in \mathbb{R}^{1 \times d_a}$ in additive compatibility function (Eq.(\ref{eq:comp_functions_2})) with a matrix $W\in\mathbb{R}^{d_e \times d_a}$, and thus produces $d_e$ scores to describe the attention of $q$ to $x_i$. }

\subsection{Self-Attention Mechanism} \label{sec:selfATT}

Self-attention mechanism is a special case of attention mechanisms, where the query $q$ stems from the input sequence itself. Self-attention mechanisms can be classified into token2token or source2token self-attention mechanism according to the type of dependency each aims to model.

\textbf{A)} \textbf{Token2token self-attention} mechanism \cite{vaswani2017attention,shen2017disan} aims at producing a context-aware representation for each token in light of its syntactic dependencies on other tokens from the same sequence. Two examples of token2token self-attention are 1) scaled dot-product self-attention which composes the multi-head self-attention \cite{vaswani2017attention}, and 2) masked self-attention used in directional self-attention \cite{shen2017disan}. 

\textbf{A.1}) \textbf{Scaled dot-product attention} mechanism \cite{vaswani2017attention} in general form has three arguments: query tokens $\bm{q} \in\mathbb{R}^{d_i \times m}$, key tokens $\bm{k} \in\mathbb{R}^{d_i \times n}$ and value tokens $\bm{v} \in\mathbb{R}^{d_h \times n}$ associated with the key tokens. It uses a scaled dot-product function to model the relationship between each query and key, and finally outputs a sequence $\bm{s}=[s_1, \dots, s_m] \in\mathbb{R}^{d_h \times m}$ such that
\begin{equation}
\bm{s} \! = \! \dpattn(\bm{q}, \bm{k}, \bm{v}) \triangleq \bm{v} \softmax(\dfrac{\bm{q}^T \bm{k}}{\sqrt{d_q}})^T  
\end{equation}
A special case of this mechanism is that the three input arguments are derived from the same source, i.e., $\bm{q}\text{/}\bm{k}\text{/}\bm{v}=f^{q\text{/}k\text{/}v}(\bm{x})$, which can be referred to as a token2token self-attention, namely scaled dot-product self-attention. As for \textbf{multi-head attention} mechanism, the input is projected into multiple subspaces, then parameter-untied scaled dot-product attention is applied to the embeddings in each subspace. The results for multiple subspaces are concatenated to form the final output $\bm{s}$, i.e., 
\begin{align}
	&\bm{s} = W^{(o)}[H_1;\dots;H_h], \\
	\notag &\text{where }H_c = \dpattn(W_c^q\bm{q}, W_c^k\bm{k}, W_c^v\bm{v}).
\end{align}
\textbf{A.2)} \textbf{Masked self-attention} mechanism \cite{shen2017disan} uses multi-dim compatibility function to model the dependency between every two tokens in a sequence, and uses positional mask to encode sequential information. It overcomes inherent problem appearing in self-attention compared to RNNs on the lack of sequential information. Its compatibility function is defined as
\begin{equation}\label{equ:maskedSA}
f(x_i, x_j) \!=\! c \cdot \tanh \{ (W^{(m)}[x_i; x_j] + b^{(m)}) / c \} \!+\! M_{i\!,j}
\end{equation}
where $c$ is a constant scalar, $W^{(m)} \in \mathbb{R}^{d_e\times 2d_e}$ is learnable weight matrix, and $M$ is a positional mask with each entry $M_{i,j} \in \{-\infty, 0\}$. When $M_{i,j}=-\infty$, applying $\softmax$ function to the alignment scores results in a zero attention probability, which cuts off the attention of $x_j$ to $x_i$. Hence, masked self-attention with an asymmetric mask, where $M_{ij}\neq M_{ji}$, can encode sequential or other  structural information \cite{shen2017disan,im2017distance}. To this end, two positional masks have been proposed to encode the forward and backward order information respectively, i.e.,
\begin{equation}\notag \label{eq:fw_bw_masks} 
M_{i,j}^{fw}\!\! =\!\! \left\{
\begin{array}{ll}
\!\!\!\!0,& \!\!\!\!\!i < j\\
\!\!\!\!-\! \infty,& \!\!\!\!\!\text{otherwise}\\
\end{array}
\right.
\hspace{-0.1em}
M_{i,j}^{bw} \!\!=\!\!\left\{\begin{array}{ll}
\!\!\!\!0, & \!\!\!\!\!i > j \\
\!\!\!\!- \infty, &\!\!\!\!\! \text{otherwise}\\
\end{array}\right.
\end{equation}
Furthermore, \textbf{directional self-attention} (DiSA) \cite{shen2017disan} concatenates the features produced by masked self-attention mechanisms with the forward and backward positional masks (i.e.,  $M^{fw}, M^{bw}$), leading to context-ware representations with bi-directional information encoded.

\textbf{B)} \textbf{Source2token self-attention} mechanism \cite{liu2016learning,lin2017structured,shen2017disan} is designed for sentence  embedding or sequence compression, which is based on the importance of each token $x_i$ to the entire source sequence $\bm{x}$ for a specific task. Specifically, it removes the query $q$ from the compatibility function $f(x_i, q)$ when computing the alignment score. For example, the compatibility function of additive source2token self-attention mechanism is to simply remove $q$ from Eq.(\ref{eq:comp_functions_2}).

\section{Proposed Models}

In this section, we firstly elaborate on tensorized self-attention (TSA) in Section \ref{sec:tsa}, which captures both pairwise and global dependencies by combining the two types of self-attention mechanisms introduced in Section~\ref{sec:selfATT}. Then, we extend TSA to multi-mask tensorized self-attention (MTSA) in Section~\ref{sec:mtsa} by applying different positional masks to TSA in multiple subspaces (multi-head fashion). Lastly, in Section~\ref{sec:model_acce}, we present an efficient computation scheme for MTSA without any high-rank tensor computation involved even if tensorized alignment scores are used.

\subsection{Tensorized Self-Attention (TSA)}\label{sec:tsa}

\begin{figure}[htbp]
	\centering
	\includegraphics[width=0.475\textwidth]{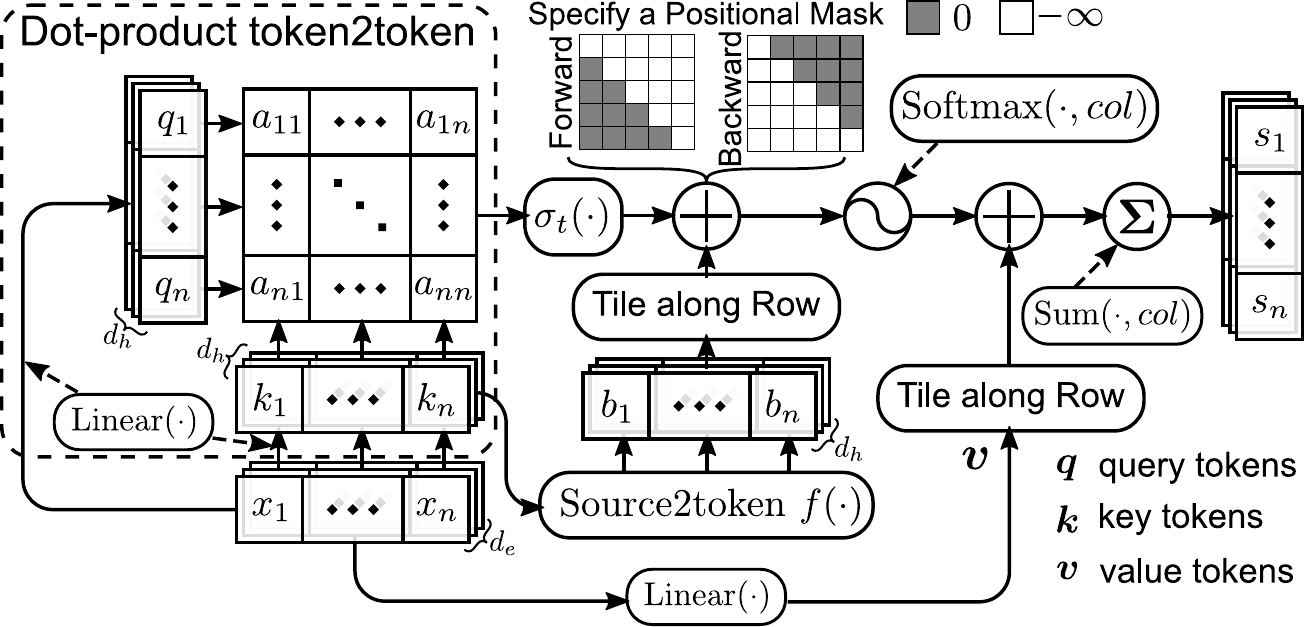}
	\caption{Tensorized self-attention (TSA) Mechanism.}
	\label{fig:tsa_network} 
	\centering
\end{figure}

Tensorized self-attention (TSA), whose structure is illustrated in Figure \ref{fig:tsa_network}, is a neural mechanism that can be trained to model both pairwise and global dependencies, while any previous self-attention mechanism only focuses on one type of dependencies. TSA models both types by combining the aforementioned token2token and source2token self-attention mechanisms. This generates an $n\times n\times d_h$ tensor containing the alignment scores between every two tokens on each feature dimension. These scores are then normalized and transformed into probability weights, which are used to sum all dependent tokens and then generate the contextual embedding for each input token. We will demonstrate later in Section \ref{sec:model_acce} that only matrix rather than tensor operation is required when executing the procedures above. 

To facilitate the elaboration of proposed models and keep the consistent notation with prior attention mechanisms, TSA first projects the input embeddings $\bm{x}$ into three spaces to represent the query, key and value tokens, respectively. 
\begin{equation}
\bm{q} \!=\! W^{(t1)}\bm{x},~\bm{k} \!=\! W^{(t2)}\bm{x},~\text{and}~~\bm{v} \!=\! W^{(t3)} \bm{x},
\end{equation}
where $W^{(t1)}, W^{(t2)} \in \mathbb{R}^{d_i \times d_e}$ and $W^{(t3)} \!\in\!\mathbb{R}^{d_h \times d_e}$ are learnable weights for projections.

TSA then integrates two kinds of compatibility functions from two self-attention mechanisms respectively. Firstly, the scaled dot-product self-attention is used to capture dependency between every two tokens. Dot-product operations are fast, and sufficient to model the \textbf{pairwise dependency} in most tasks. Its compatibility function is
\begin{align}
	&f^t(k_i, q_j) =  \langle k_i,~q_j \rangle / \sqrt{d_i},~~\forall i,j \in[n],
\end{align}
where $\langle \cdot ,\cdot \rangle$ is inner-product operation. Then, a multi-dim source2token self-attention mechanism is used to estimate the contribution of each token to the given task on each feature dimension. It aims at capturing the importance of each token to the entire input sequence w.r.t. the task, i.e., the \textbf{global dependency}. The multi-dim extension only linearly increases the memory and computation of source2token self-attention by a multiplicative factor $d_h$, but is essentially helpful to improve expressive capability in line with prior works \cite{shen2017disan}. Its compatibility function is
\begin{equation}
f^s(k_i) = W^{(s2)}\sigma_m (W^{(s1)}k_i + b^{(s1)}) + b^{(s2)},
\end{equation}
where $\forall i \in[n]$, $W^{(s1)} \!\in\!\mathbb{R}^{d_a \times d_i}, W^{(s2)} \!\in\!\mathbb{R}^{d_h \times d_a}$ are the learnable weights, and $\sigma_m (\cdot)$ is an activation function. 
The compatibility function used in TSA broadcasts the scalar alignment score $f^t(k_i, q_j) \in\mathbb{R}$ computed by the token2token self-attention to all $d_h$ feature dimensions, and then adds them to the feature-wise score vector $f^s(k_i) \in\mathbb{R}^{d_h}$ computed by the source2token self-attention. In addition, the positional masks from masked self-attention (in Section~\ref{sec:selfATT}) are also integrated to encode sequential and structural information. These yield following compatibility function of TSA. 
\begin{align}
\left[f^{tsa}(k_i, q_j)\right]_l  =  ~~~~~&\label{equ:tsa_comp_func}\\ 
\notag \sigma_t  \left(f^t(k_i, q_j)\right) &+ \sigma_s(\left[f^s(k_i)\right]_l) + M_{i,j},
\end{align}
where $\forall i,j \in[n],~\forall l\in[d_h]$. $ \sigma_t (\cdot)$ and $ \sigma_t (\cdot)$ are two scale functions. They control the way to combine two kinds of scores and their weights, more details of which are elaborated in Appendix A.1.
We also show heatmaps of the token2token and source2token alignment scores in Appendix E.

For each query token $q_j$, a $\softmax$ function is applied to the alignment scores $[f^{tsa}(k_i, q_j)]_{i=1}^{n}$ on each feature dimension, resulting in a categorical distribution over all value tokens $[v_i]_{i=1}^{n}$ based on corresponding key tokens $[k_i]_{i=1}^{n}$. The probability of token $q_j$ attending to $v_i$ on the $l^{th}$ feature dimension (i.e., $z_l=i$) is
\begin{align}\label{equ:TSAprop}
	p(z_l\!=\!i|\bm{k}, \!q_j) \!\triangleq\! [p_i^j]_l \triangleq \dfrac{e^{\left[f^{tsa}(k_i, q_j)\right]_l}}{\sum\nolimits_{g=1}^{n}e^{\left[f^{tsa}(k_g, q_j)\right]_l}}\!,\!
\end{align}
where, $\forall i,j \in[n],~\forall l\in[d_h]$. TSA outputs a contextual embedding for each input token on every feature dimension as the weighted sum of all the value token embeddings on that dimension, where the weights are provided by the probabilities in Eq.(\ref{equ:TSAprop}). It is the expectation of sampling a value token embeddings on each feature dimension according to the feature-wise probability distribution, i.e., 
\begin{align}\label{equ:TSAoutput}
	&\bm{s}\triangleq [s_j]_{j=1}^n,~~\text{where}~\\
	\notag &s_j\triangleq \left[\mathbb{E}_{i\sim p(z_l|\bm{k}, q_j)}([v_i]_l)\right]_{l=1}^{d_h} = \sum\nolimits_{i=1}^{n}p_i^j \cdot v_i
\end{align}

\subsection{Multi-Mask Tensorized Self-Attention (MTSA) Mechanism}\label{sec:mtsa}

Rather than computing attention in the original input space, multi-head attention \cite{vaswani2017attention} projects the input sequence to multiple subspaces, applies attention to the projected embedding in each subspace, and concatenates their outputs at last. The computations associated with multiple heads can be completed in parallel. By using adequate heads, each with a low-dimensional subspace (i.e., the representation dimension for each head is updated by $d_h \! \gets\! d_h / h$ where $h$ is the number of head), it reduces parameters and memory/computation cost and increases diversity of the attention. 
In addition, to encode different kinds of sequential or structural information, multiple different positional masks (e.g., forward, backward and multi-length window) can be further applied to the multiple heads.  

The memory-/time-efficiency and expressive power of TSA can be improved by using the combination of the multi-head and multi-mask techniques introduced above. By writing TSA mechanism as a function $\tsa (\bm{x}, M)$ with input sequence $\bm{x} \in \mathbb{R} ^{d_e \times n}$ and a positional mask $M \in \mathbb{R} ^{n \times n}$, and the output given by Eq.(\ref{equ:TSAoutput}), multi-mask tensorized self-attention (MTSA) produces
\begin{align} 
\bm{s} =& W^{(o)}[H_1; \dots; H_h], \label{eq:mtsa_output} \\
\notag  &\text{where}~H_c = \tsa\nolimits^c(\bm{x}, M^{c}),
\end{align}
where $W^{(o)}\in\mathbb{R}^{h\cdot d_h \times h\cdot d_h}$,
$h$ is the number of heads, 
$\tsa\nolimits^c$ denotes the $c^{th}$ parameter-independent TSA block that produces a $d_h$-dim representation in the $c^{th}$ subspace,
$M^{c}$ represents the positional mask applied to attention in the $c^{th}$ subspace, $[\cdot; \dots;\cdot]$ denotes a vertical concatenation operation, and $\bm{s}\in\mathbb{R}^{h\cdot d_h \times n}$ is the output of MTSA. In our experiments, we apply forward mask to half of the heads and apply backward mask to the other half to encode bi-directional order information of the input sequence.

\subsection{Computation-Optimized MTSA}\label{sec:model_acce}

\begin{algorithm} [htbp] \small
	\caption{\small Multi-Mask Tensorized Self-Attention}\label{alg:mtsa_optimized}
	\begin{flushleft}
		\textbf{Input:} input sequence $\bm{x}\in\mathbb{R}^{d_e \times n}$, head number $h$, subspace dimension $d_h$, positional masks $\{M^{c}\}_{c=1}^{h}$, and weights/biases: 
		$\{W_c^{(t1)}, W_c^{(t2)} \in\mathbb{R}^{d_i\times d_e}, W_c^{(t3)} \in\mathbb{R}^{d_h\times d_e}, W_c^{(s1)} \in\mathbb{R}^{d_a\times d_i}, W_c^{(s2)} \in\mathbb{R}^{d_h\times d_a}, 
		\text{and}~b_c^{(s1)},~b_c^{(s2)}\}_{c=1}^{h},~\text{and}~W^{(o)}$\\
		\textbf{Output:} contextual embeddings  $\bm{s}\!=\![s_1, \!\dots\!, s_n] \!\! \in \!\! \mathbb{R}^{h\cdot d_h \!\times\! n}$
	\end{flushleft}
	\begin{algorithmic}[1]
		\ForAll{$c=1, \dots, h$}  \Comment Computing $h$-head in parallel
		\State $\bm{q^c},~\bm{k^c},~\bm{v^c}  \gets W_c^{(t1)}\bm{x},~W_c^{(t2)}\bm{x},~W_c^{(t3)} \bm{x}$   
		\State $R_c \! \gets \! \dfrac{(\bm{k^c})\!^T  \bm{q^c}}{\sqrt{d_h}} $ \label{alg:ln:t2t_score} \Comment $n\!\times\! n$ token2token attention scores
		\State $S_c \gets W_c^{(s2)}\sigma_m (W_c^{(s1)} \bm{k^c} + b_c^{(s1)}) + b_c^{(s2)}$ 
		$~~~~~~~~~~~~~~~~~~~~~~$ 
		\label{alg:ln:s2t_score}  \Comment $d_h\times n$ scores of source2token attention
		\State $E_c^R \gets \exp(\sigma_t(R_c)) \cdot \exp(M^c)$ 
		$~~~~~~~~~~~~~~~~~~~~~~~~~~~~~~~~~~~~~~~~~~$ 
		\label{alg:ln:apply_mask}\Comment Applying mask $M^c$ to token2token weights
		\State $E_c^S \gets \exp(\sigma_s(S_c));~~E_c^X \gets\bm{v^c}\cdot E_c^S$  
		$~~~~~~~~~~~~~~~~~~~~~~~~~$  
		\label{alg:ln:apply_x}  \Comment Applying source2token weights $E_c^S$ to $\bm{v^c}$
		\State $H_c \gets E_c^X E_c^R  / E_c^S E_c^R $ 
		\label{alg:ln:wei_sum} \Comment Applying masked token2token weights $E_c^R$ and normalizing
		\EndFor
		\State \textbf{Return} $\bm{s} \gets W^{(o)}[H_1; \dots; H_h]$ 
		$~~~~~~~~~~~~~~~~~~~~~~~~~~~~~~~~~~~~~~~~~~~~~~~~$ 
		\label{alg:placeholder3}\Comment Vertical concatenation of the outputs from all $h$ heads
	\end{algorithmic}
\end{algorithm}

As shown in Eq.(\ref{equ:tsa_comp_func}) and Eq.(\ref{equ:TSAprop}), TSA or each head of MTSA needs to compute the attention scores and probabilities as $n\times n\times d_h$ tensors. In accordance with multi-dim self-attention \cite{shen2017disan}, this makes TSA more expressively powerful and improves the final performance for sequence modeling, but terribly leads to memory explosion and computational bottleneck on long sequences with large $n$ and $d_h$. Fortunately, in MTSA, it is possible to significantly reduce the demand on computations to matrix-only operations by exploring the computational structure. 

A memory-optimized and highly-parallelizable computation scheme for MTSA is given in Algorithm \ref{alg:mtsa_optimized}. For each head, the score matrices of token2token and source2token are computed in steps \ref{alg:ln:t2t_score} and \ref{alg:ln:s2t_score} respectively. Then, we combine token2token scores with the positional mask to form a new mask in step \ref{alg:ln:apply_mask}, and compute the $d_h\times n$ output embedding with the weighs from the multi-dim source2token self-attention in step \ref{alg:ln:apply_x}. Finally, in step \ref{alg:ln:wei_sum}, we apply the new mask from step \ref{alg:ln:apply_mask} to the weighted embedding from step \ref{alg:ln:apply_x} and complete the normalization. This procedure generates the exactly same output as Eq.(\ref{eq:mtsa_output}) (as rigorously proved in Appendix A.2) but no any tensor operation is incurred. More details about memory-efficiency, time-efficiency and attention dropout are elaborated in Appendix A.3.

\section{Experiments} \label{sec:experiments}

We compare MTSA with commonly-used context fusion baselines on several NLP tasks\footnote{Codes for Experiments are released at \url{https://github.com/taoshen58/msta}.}. When addressing a sentence embedding problem, a multi-dim source2token self-attention is applied on the top of context fusion module to produce the sequence embedding. Codes are implemented in Python with Tensorflow and executed on a single NVIDIA GTX 1080Ti graphics card. In addition, data for both time cost and memory consumption are collected under Tensorflow-1.7 with CUDA9 and cuDNN7. The \textbf{fair} and \textbf{reliable} experiment setups are elaborated in Appendix B. 

\begin{table*}[htbp]\footnotesize
	\centering
	\setlength{\tabcolsep}{3pt}
	\begin{tabular}{@{}lcccccc@{}} 
		\toprule
		\textbf{Model}& \textbf{$\boldsymbol{|\theta|}$} & \textbf{Time/Epoch}&\textbf{Inf. Time}&\textbf{Memory}&\textbf{Train Acc.} & \textbf{Test Acc.} \\ 
		\midrule
		300D SPINN-PI encoders \cite{bowman2016fast}&            3.7m&                 &&&              89.2&               83.2\\ 
		600D Bi-LSTM encoders \cite{liu2016learning}&            2.0m&                  &&&           86.4&               83.3\\ 
		600D Bi-LSTM enc.+intra-attn \cite{liu2016learning}&            2.8m&             &&&                 84.5&               84.2\\ 
		600D Deep Gated Attn. \cite{chen2017recurrent}&           11.6m&                 &&&              90.5&               85.5\\
		600D Gumbel TreeLSTM enc. \cite{choi2017learning}&            10.0m&              &&&                 93.1&               86.0\\
		600D Residual stacked enc. \cite{nie2017shortcut}&  29.0m	&  &&&  	91.0 &	86.0\\
		300D Reinforced SAN \cite{shen2018reinforced}&           3.1m&              404s&&&                 92.6	&               86.3\\
		Distance-based SAN \cite{im2017distance}&           4.7m&              416s&&&                 89.6&               86.3\\
		\midrule
		Bi-LSTM \cite{graves2013hybrid}&            2.9m&            854s&9.1s&942MB&                90.4&             85.0\\ 
		Bi-GRU \cite{chung2014empirical}&            2.5m&            850s&9.4s&810MB&          91.9&             84.9\\ 
		Multi-CNN \cite{kim2014convolutional}&            1.4m&           137s&1.4s&208MB&             89.3&              83.2\\ 
		Hrchy-CNN \cite{gehring2017convolutional}&            3.4m&           195s&1.8s&309MB&             91.3&              83.9\\ 
		Multi-head \cite{vaswani2017attention}&            2.0m&                   179s&1.5s&466MB&            89.6&               84.2\\ 
		DiSA \cite{shen2017disan}&            2.3m&                390s&5.2s&6682MB&          91.1&               85.6\\ 
		Bi-BloSA \cite{shen2018biblosan}&            4.1m&                303s&3.2s&1600MB&          91.6&               85.8\\ \midrule
		MTSA &            2.9m&              180s&1.6s&558MB&       91.8&               \textbf{86.3}\\ \bottomrule
	\end{tabular}
	\caption{ Experimental results for different methods with comparative parameter number on SNLI. 
		\textbf{$\boldsymbol{|\theta|}$}: the number of parameters (excluding word embedding part); 
		\textbf{Time/Epoch}: averaged training time per epoch with batch size 128; 
		\textbf{Inf. Time}: averaged dev inference time with batch size 128;
		\textbf{Memory}: memory load on synthetic data of sequence length 64 and batch size 64 with back-propagation considered;
		\textbf{Train Acc.} and \textbf{Test Acc.}: the accuracies on training/test sets. All state-of-the-art methods in leaderboard are listed in Table \ref{table:exps_snli}\&\ref{table:exps_transfer} up to Sep. 2018.}
	\label{table:exps_snli}
\end{table*}

The context fusion baselines include
\textbf{1) Bi-LSTM} \cite{graves2013hybrid}: 600D bi-directional LSTM consisting of 300D forward plus 300D backward LSTMs, 
\textbf{2) Bi-GRU} \cite{chung2014empirical}: 600D bi-directional GRU, 
\textbf{3) Multi-CNN} \cite{kim2014convolutional}: three CNNs with 200D kernels to model 3/4/5-grams respectively, 
\textbf{4) Hrchy-CNN} \cite{gehring2017convolutional}: 3-layer 300D stacked CNN with kernel size 5, gated linear units \cite{dauphin2016language} and residual connections \cite{he2016deep},
\textbf{5) Multi-head} \cite{vaswani2017attention}: 600D multi-head self-attention with 8 heads (75-dim subspace per head) and positional embedding used by \citeauthor{vaswani2017attention}~\shortcite{vaswani2017attention},
\textbf{6) DiSA} \cite{shen2017disan}: 600D directional self-attention mechanism consisting of 300D forward and 300D backward masked self-attentions, and
\textbf{7) Bi-BloSA} \cite{shen2018biblosan}: 600D bi-directional block self-attention with intra-/inter-block self-attention, aiming to reduce the time and space complexities of multi-dim self-attention by using hierarchical structure.

\subsection{Natural Language Inference} \label{sec:nli}

Natural language inference (NLI) aims at speculating on the relationship between a premise and a corresponding hypothesis, where the relationship could be \textit{entailment}, \textit{neutral} or \textit{contradiction}. In experiments, we first compare MTSA with other baselines on the Stanford Natural Language Inference \cite{bowman2015snli}  (SNLI) dataset. 

Following the method of applying \emph{sentence-encoding} model to NLI given by \citeauthor{bowman2016fast}~\shortcite{bowman2016fast}, two parameter-tied sentence-encoding models are used to generate embeddings for premise and hypothesis, resulting in $s^p$ and $s^h$ respectively. The concatenation of $s^p$, $s^h$, $s^p - s^h$ and $s^p \odot s^h$ representing the relationship is passed into a 3-way neural classifier for final prediction.

The experimental results of the models from the official leaderboard, baselines, and MTSA are shown in Table \ref{table:exps_snli}. MTSA achieves state-of-the-art performance with less time and memory cost. Compared to the methods from the leaderboard, MTSA outperforms RNN-based encoders (e.g., Residual stacked enc.), RNN+attention encoders (e.g., Deep Gated Attn.) and even parsing trees based encoders (e.g., Gumbel TreeLSTM enc.) by a large margin. Compared to the two competitive self-attention networks with complicated and expensive training computations, MTSA trained in end-to-end manner achieves the same state-of-the-art performance by using much fewer parameters and less computational time.

Compared to baselines, MTSA is $4\!\sim\!5\!\times$ faster than RNN-based models and outperforms CNN-based models given a similar number of parameters and computation time. Moreover, compared to the dot-product self-attention (Multi-head), MTSA costs similar time and memory but performs more expressively powerful self-attention, and thus achieves better performance. Furthermore, compared to the multi-dim self-attention (DiSA and Bi-BloSA), MTSA uses much less memory and time but even produces much better prediction quality.

\begin{table}[t] \small
	\centering
	\setlength{\tabcolsep}{4pt}
	\begin{tabular}{@{}lcccc@{}}
		\toprule
		\multirow{2}{*}{\textbf{Model}} &\multicolumn{2}{c}{\textbf{SNLI}}&\multicolumn{2}{c}{\textbf{MultiNLI}}\\ \cmidrule(l){2-5}
		& Dev& Test& Match&  Mismatch\\ \midrule 
		BiLSTM w/ Shortcut$^a$& --& 86.0& \underline{74.6}& 73.6\\ 
		BiLSTM w/ Gen-Pooling$^b$ & --& \underline{86.6}& 73.8& \underline{74.0}\\  
		HBMP$^c$& --& \underline{86.6}& 73.7& 73.0\\  
		\midrule
		Transfer + Multi-Head  &            86.9&          86.6&      76.3&   75.7\\
		Transfer + MTSA   &  \textbf{87.2}&   \textbf{86.9} &  \textbf{76.7}&   \textbf{76.4}\\
		\bottomrule
	\end{tabular}
	\caption{ Experimental results on \textit{sentence-encoding} based SNLI and MultiNLI benchmark tasks. ``\textbf{Transfer}'' denotes pretrained language model on large corpus for transfer learning, which detailed by \citeauthor{radford2018improving}~\shortcite{radford2018improving}. References: $^a$\cite{nie2017shortcut}, $^b$\cite{chen2018enhancing}, $^c$\cite{talman2018natural}.}
	\label{table:exps_transfer}
\end{table}

In addition, to further improve the state-of-the-art performance, in contrast to training from scratch, a language model built on the Transformer \cite{vaswani2017attention} unsupervisedly pretrained on large English corpus (detailed by \citeauthor{radford2018improving}~\shortcite{radford2018improving}) is transfered for the baseline and proposed models for \textit{sentence-encoding} based NLI tasks. As shown in Table~\ref{table:exps_transfer}, MTSA integrated with pretrained language model can achieve new state-of-the-art accuracy on both SNLI and Multi-Genre Natural Language Inference (MultiNLI) \cite{williams2017broad}\footnote{All test results are Evaluated on Kaggle official websites: https://www.kaggle.com/c/multinli-matched-open-evaluation and https://www.kaggle.com/c/multinli-mismatched-open-evaluation} among all \textit{sentence-encoding} models. 

An ablation study of MTSA is shown in Table \ref{table:exps_snli_ablation} to verify the capability of its each part in context fusion. The results show that token2token (modeling pairwise dependency), source2token (modeling global dependency), and positional masks (encoding sequential information) all contribute important information to sequence modeling, and the contributions are complementary. 

\begin{table}[t] \small
	\centering
	\setlength{\tabcolsep}{1.9pt}
	\begin{tabular}{@{}lccc@{}}
		\toprule
		\textbf{Model}& \textbf{$\boldsymbol{|\theta|}$} &\textbf{Inf. Time}&\textbf{Test Acc.}\\ \midrule
		MTSA  &            2.9m&        1.6&       86.3\\ 
		MTSA w/o fw\&bw masks  &            2.9m&          1.6&      85.3 (-1.0)\\
		MTSA w/o token2token  &            2.5m&          1.5&      85.8 (-0.5)\\
		MTSA w/o source2token& 2.5m&      1.4&       84.9 (-1.4)\\
		MTSA w/o proposed modules& 1.8m&     1.1&        84.3 (-2.0)\\
		\bottomrule
	\end{tabular}
	\caption{ An ablation study of MTSA on SNLI.}
	\label{table:exps_snli_ablation}
\end{table}

\begin{table*}[t] \small
	\centering
	\setlength{\tabcolsep}{3pt}
	\begin{tabular}{@{}lccccccccccccc@{}}
		\toprule
		\multirow{2}{*}{\textbf{Models}}&\textbf{Training} & \multicolumn{4}{c}{\textbf{Development}} & \multicolumn{4}{c}{\textbf{WSJ Test}} & \multicolumn{4}{c}{\textbf{Brown Test}} \\	\cmidrule(l){3-14} 
		&\textbf{Time}& \textbf{P}&\textbf{R}&\textbf{F1}&\textbf{Comp.}&\textbf{P}&\textbf{R}&\textbf{F1}&\textbf{Comp.}&\textbf{P}&\textbf{R}&\textbf{F1}&\textbf{Comp.} \\ 
		\midrule
		\citeauthor{tackstrom2015efficient}~\shortcite{tackstrom2015efficient}&&    81.2&76.2&78.6&54.4&    82.3&77.6&79.9&56.0&    \textbf{74.3}&68.6&71.3&39.8 \\
		\citeauthor{zhou2015end}~\shortcite{zhou2015end}&&     79.7&79.4&79.6&-&    82.9&82.8&82.8&-&    70.7&68.2&69.4&-\\
		\citeauthor{he2017deep}~\shortcite{he2017deep}&& 81.6& 81.6& 81.6&62.3&    83.1& 83.0& 83.1&64.3&    72.8&71.4 &72.1&44.8\\
		\citeauthor{he2018jointly}~\shortcite{he2018jointly}&& -& -& -&-&    -& -& 83.9&-&    -&- &73.7&-\\
		\citeauthor{strubell2018linguistically}~\shortcite{strubell2018linguistically}&& -& -& -&-&    \textbf{84.7}& 84.2& 84.5&-&    73.9&72.4 &73.1&-\\
		\midrule 
		Bi-LSTM \cite{graves2013hybrid}&72h& 81.8& 83.4& 82.6& 63.3&    83.0& 84.0& 83.5& 64.6&    72.3& 72.8& 72.5& 46.8\\
		Multi-CNN \cite{kim2014convolutional}&19h& 75.2& 79.6& 77.3& 53.6&    77.3& 80.9& 79.0& 55.5&    68.3& 70.3& 69.3& 41.9\\
		Multi-head$^*$ \cite{tan2017deep}&20h& 82.6& 83.6& 83.1&65.2&    84.5& 85.2& \textbf{84.8}&66.4&    73.5&\textbf{74.6} &74.1&48.4\\
		\midrule 
		MTSA&20h& \textbf{82.8}& \textbf{84.4}&\textbf{83.6}& \textbf{65.4}&    84.2& \textbf{85.3}& \textbf{84.8}& \textbf{67.0}&    \textbf{74.3}& \textbf{74.6}& \textbf{74.5}& \textbf{49.1}\\
		\bottomrule
	\end{tabular}
	\caption{Experimental Results of SRL for single models on CoNLL-05 with gold predicates. $^*$Multi-head baseline is equivalent to the model in \citeauthor{tan2017deep}~\shortcite{tan2017deep}. For fair comparisons, first, we use  the hyper-parameters provided by \citeauthor{tan2017deep}~\shortcite{tan2017deep} instead of tuning them; second, all listed models are independent of external linguistics information, e.g., PoS, dependency parsing.}
	\label{table:exps_srl_new}
\end{table*}

\subsection{Semantic Role Labeling}

To verify the capability of MTSA in generating context-aware representation of each token, we compare it with baselines on semantic role labeling (SRL) task, which aims to tag each token from an input sequence with a label for its semantic role. Particularly, given a sentence, the goal of SRL is to identify the arguments of each target verb into semantic roles, which can benefit many downstream NLP tasks. SRL has two steps: 1) assigning either a semantic argument or non-argument to a given predicate and 2) labeling a specific semantic role for the identified argument.

We follow the experimental setup in \citeauthor{tan2017deep}~\shortcite{tan2017deep}, where the SRL task is treated as a BIO tagging problem. \citeauthor{tan2017deep}~\shortcite{tan2017deep} designed a deep attentive neural net by stacking multi-head self-attention, named as deepatt, to perform context fusion, whose output is then passed to a neural classifier to make the final decision. The results achieved by previous methods, baselines, and MTSA are shown in Table \ref{table:exps_srl_new}, which demonstrates that MTSA achieves new state-of-the-art performance on the CoNLL-05 dataset by costing similar training time as CNN and multi-head self-attention baselines.

\subsection{Sentence Classifications}

\begin{table}[t]  \footnotesize %
	\centering
	\setlength{\tabcolsep}{0.3pt}
	\begin{tabular}{@{}lccccc@{}}
		\toprule
		\textbf{Model}& \textbf{CR} & \textbf{MPQA} & \textbf{SUBJ}& \textbf{TREC}&\textbf{SST-5}\\ \midrule 
		cBoW$^a$&	79.9& 86.4& 91.3& 87.3&  /\\
		Skip-thought$^b$&	81.3& 87.5& 93.6& 92.2&/\\
		DCNN$^c$&	/& /& /& 93.0&  48.5\\
		SRU$^d$&	84.8(1.3)& 89.7(1.1)& 93.4(0.8)& 93.9(0.6)& /\\
		CNNs$^d$&	82.2(.2)& 88.8(1.2)& 92.9(0.7)& 93.2(0.5)&  /\\\midrule
		Bi-LSTM &  84.6(1.6)&  90.2(0.9)&  \textbf{94.7(0.7)}&  94.4(0.3)&   49.9(0.8)\\
		Multi-head & 82.6(1.9)&  89.8(1.2)& 94.0(0.8)&  93.4(0.4)&    48.2(0.6)\\
		DiSA &	 84.8(2.0)&  90.1(0.4)&  94.2(0.6)&  94.2(0.1)&  51.0(0.7)\\ 
		Bi-BloSA &	84.8(0.9)& 90.4(0.8)& 94.5(0.5)& 94.8(0.2)&   50.6(0.5)\\ 
		\midrule
		MTSA &	\textbf{84.9(2.4)}& \textbf{90.5(0.6)}& 94.5(0.6)& \textbf{95.3(0.3)}& \textbf{51.3(0.7)}\\ 
		\bottomrule
	\end{tabular}
	\caption{ Experimental results on five sentence classification benchmarks. References: $^a$\cite{mikolov2013efficient}, $^b$\cite{kiros2015skip}, $^c$\cite{kalchbrenner2014convolutional}, $^d$\cite{lei2017sru}.}
	\label{table:exps_sc_accu}
\end{table}

The goal of sentence classification is to predict the correct label for a sentence in various scenarios. We evaluate the models on five sentence classification benchmarks for different NLP tasks, which include 
\textbf{1) CR} \cite{hu2004mining}: customer reviews  of various products to predict whether the review is  positive or negative, 
\textbf{2) MPQA} \cite{wiebe2005annotating}: an opinion polarity detection subtask of the MPQA dataset, 
\textbf{3) SUBJ}  \cite{pang2004sentimental}: subjectivity dataset where a label indicates whether a sentence is subjective or objective, 
\textbf{4) TREC} \cite{li2002learning}: question-type classification dataset which classifies the question sentences into six classes, 
\textbf{5) SST-5} \cite{socher2013recursive}: the Stanford Sentiment Treebank dataset with five sentiment labels.
The reported accuracies for CR, MPQA, and SUBJ are the mean of 10-fold cross validation. The accuracies for TREC are the mean of five runs on the dev set, and the accuracies for SST-5 are the mean of five runs on the test set. All standard deviations are shown in parentheses.

The prediction accuracies achieved on these five benchmarks are shown in Table \ref{table:exps_sc_accu}. MTSA achieves the best prediction accuracy on CR, MPQA, TREC and SST-5 benchmarks with better time efficiency and a lower memory load. 

\subsection{Machine Translation}

We also evaluate proposed model on WMT 2014 English-German translation task for exhaustive comparisons with multi-head attention. We replace multi-head self-attention modules in the encoder of official Transformer implementation with MTSA module and do not tune the hyperparameters. Although our computation resources is limited, we use two training setups and also introduce \emph{t}-test to ensure that MTSA consistently outperforms multi-head self-attention in Transformer. 

For \textbf{Setup1}, we use default hyperparameter set of \emph{transformer\_base\_single\_gpu} provided by official implementation with $1\times\text{P100}$ , batch size of 2048 and training step of 250K, and report BLEU value for the last checkpoint. For \textbf{Setup2}, we use the hyperparameter set of \emph{transformer\_base} with the modification of 1) using $4\times$ instead of $8\times\text{P100}$, 2) increasing batch size from 4096 to 6144 per GPU, and 3) using training step of 133K. More details of the training setups for translation task are described in Appendix B.1.

\begin{table}[t] \small
	\centering
	\begin{tabular}{lcc}
		\toprule
		\textbf{Model}                   & \textbf{Multi-head (Transformer)} & \textbf{MTSA} \\ \midrule
		\textbf{Param\#}                   & 61.38M                  & 61.58M            \\ \midrule
		\multirow{2}{*}{\textbf{Setup1}} & 23.64               & 24.09         \\ \cmidrule(l){2-3} 
		& \multicolumn{2}{c}{\emph{p-value:} 0.001 (6 runs)}    \\ \midrule
		\multirow{2}{*}{\textbf{Setup2}} & 26.98               &  27.21         \\ \cmidrule(l){2-3} 
		& \multicolumn{2}{c}{\emph{p-value:} 0.080 (3 runs)}    \\\bottomrule
	\end{tabular}
	\caption{ Results for the Transformer with either multi-head self-attention or proposed MTSA. The reported BLEU values for Setup 1 and 2 are the mean of 5 and 3 runs respectively.}
	\label{table:exp_nmt}
\end{table}

As shown in Table \ref{table:exp_nmt}, with small p-value for both training setup 1 and 2, the encoder with MTSA significantly outperforms that with multi-head self-attention, which demonstrates that multi-dim based MTSA modeling both pairwise and global dependencies is more expressive than dot-product based multi-head self-attention. Although the results do not improve state-of-the-art BLEU value of machine translation task, the purpose of this experiment to verify the effectiveness of MTSA in contrast to dot-product based multi-head self-attention is accomplished.

\section{Conclusion}

In conclusion, MTSA is highly parallelizable with more expressive power since it efficiently captures the pairwise dependency at token level, but delicately models the global dependency at feature level, and distributes computations to multiple heads, each equipped with a distinct positional mask. These lead to a sweet spot of the trade-off between performance and efficiency,  and make MTSA as memory-efficient as CNN and scalable to long sequences but outperform previous (and even multi-dim) self-attention mechanisms in terms of prediction quality. The experiments conducted on nine NLP tasks verify that the MTSA can reach state-of-the-art performance with appealing efficiency. 

\section{Acknowledgments}
This research was funded by the Australian Government through the Australian Research Council (ARC) under grants 1) LP160100630 partnership with Australia Government Department of Health and 2) LP150100671 partnership with Australia Research Alliance for Children and Youth (ARACY) and Global Business College Australia (GBCA). We also acknowledge the support of NVIDIA Corporation and MakeMagic Australia with the donation of GPUs.

\bibliography{ref}
\bibliographystyle{acl_natbib}

\newpage
\appendix

This appendix provides the following details about ``multi-mask tensorized self-attention'' (MTSA). 
\begin{itemize}
	\item \textbf{Appendix \ref{app:supplemental}:} supplemental contents for proposed MTSA model;
	\item \textbf{Appendix \ref{app:training}:} training setups and hyper-parameter used in the experiments;
	\item \textbf{Appendix  \ref{app:related}:} related works of self-attention mechanisms and practical applications;
	\item \textbf{Appendix  \ref{app:discussion}:} discussion and future works;
	\item \textbf{Appendix  \ref{app:vis}:} a visualization of token2token and source2token alignment scores in MTSA mechanism with forward and backward positional masks.
\end{itemize}

\section{Supplemental Contents for MTSA} \label{app:supplemental}

\subsection{Scale functions}

The $\sigma_t(\cdot)$ and $\sigma_s(\cdot)$ in Eq.(10) are either parameterized or non-parameterized scale functions, which are hyperparameters of the proposed model. They can adjust the manner and weights of the combination of the two alignment score entries.

In this work, we simply focus on non-parameterized scale functions, switching between $\log(\sigmoid(\cdot))$ and $\identity(\cdot)$, which control the way to combine the two kinds of alignment scores in the attention mechanism. In particular, since the summed alignment score will be passed into a $\softmax$ function with exponential operation for attention probabilities, $\log(\sigmoid(\cdot))$ function will provide a $\sigmoid$-scaled multiplicative item for the combination during the normalization of $\softmax$, in contrast to the additive item provided by $\identity(\cdot)$. 
In addition, there are two other reasons to leverage the scale functions: 1) as stated in \citeauthor{vaswani2017attention}~\shortcite{vaswani2017attention}, and similar to $\softmax$ with temperature, $\log(\sigmoid(\cdot))$ avoids large alignment scores, which as the inputs to $\softmax$ function will result in extremely small gradient; 2) without scale function, the alignment score can be very large and may cause numerical problems during backpropagation.

\subsection{Equivalence of MTSA and Its Memory-Optimized Computation Scheme} \label{app:proof}

In this section, we rigorously prove that Algorithm 1 outputs the same results as Eq.(13). In the following analysis, we remove the subscript $c$ as the index of heads in Algorithm 1 Step 2-7 for simplicity, and use $i,~j~\text{and}~ l$ to indicate the indices of key/value tokens, query tokens and feature channels, respectively. We have that,
\begin{align}
\notag	&H_{l, j} = \sum\nolimits_{i=1}^{n}  \{E^{X}_{l,i} \cdot E^{R}_{i, j}\} / \sum\nolimits_{i=1}^{n} \{E^{S}_{l,i} \cdot E^{R}_{i, j}\}, \\
\notag &~~~~~~~~~~~~~~~~~~~~~~~~~~~~~~~~~~~\textit{// Step 7 of Algorithm 1}\\
\notag	& = \dfrac{\sum\nolimits_{i=1}^{n} \bm{v}_{l, i} \cdot \exp(S_{l, i}) \cdot \exp(\sigma_t(R_{i,j}) + M_{i,j}) }
{\sum\nolimits_{i=1}^{n} \exp(S_{l, i}) \cdot \exp(\sigma_t(R_{i,j}) + M_{i,j}) }, \\
\notag &~~~~~~~~~~~~~~~~~~~~~~~~~~~~~~~~\textit{// Step 5-6 of Algorithm 1} \\
\notag	& \!=\! \sum_{i=1}^{n}\! \! \dfrac{\bm{v}_{l,i} \!\cdot\! \exp(\sigma_s([f^s(k_i)]_l) \!\!+\!\! \sigma_t(f^t(k_i,\! q_j)) \!\!+\!\! M_{i,j}) }
{\!\sum\nolimits_{g=1}^{n}\! \exp(\sigma_s([f^s(k_g)]_l) \!\!+\!\! \sigma_t(f^t(k_g, \!q_j)) \!\!+\!\! M_{g,j}) }, \\
\notag &~~~~~~~~~~~~~~~~~~~~~~~~~~~~~~~~~~~\textit{//  Step 3-4 and Eq.(8-9)} \\
\notag	& = \sum_{i=1}^{n} [v_i]_l \!\cdot\! \dfrac{ \exp([f^{tsa}(k_i, q_j)]_l) }
{\sum\nolimits_{g=1}^{n} \exp([f^{tsa}(k_g, q_j)]_l) }, \textit{// Eq.(10)}\\
\notag	& = \sum\nolimits_{i=1}^{n} [v_i]_l\cdot [p^j_i]_l, ~~\textit{// Eq.(11-12)}\\
\notag	& = [s_j]_l = \bm{s}_{l,j}, ~~~ \forall j \in[n], \forall l\in[d_h] .
\end{align}
Hence, the computation scheme in Algorithm 1 produces the exactly same output as the original MTSA but does not require any high-rank tensor operation. Therefore, it produces the expressively powerful tensorized alignment scores but is computed as fast and as memory-efficiently as a CNN.

\subsection{More Details about Algorithm 1}

\textbf{Memory-Efficiency} (illustrated in Figure~1(a)): Compared to multi-dim token2token self-attention \cite{shen2017disan} that inherently requires 4-D tensors (with the shape of [\textit{batch size}, \textit{sequence length}, \textit{sequence length}, \textit{feature channels}]) to store the alignment scores during the training phase, MTSA does not use any high-rank tensor operation but only matrix multiplications to avoid memory explosion .

\textbf{Time-Efficiency} (illustrated in Figure~1(b)): MTSA is highly parallelizable because its computations can be distributed to multiple subspaces, and only a few matrix multiplications (which are also highly parallelizable) occur in each subspace. Compared to dot-product based multi-head attention, multi-dim based MTSA only uses extra two fully-connected layers and two element-wise matrix operations.

\textbf{Attention Dropout:} Similar to \citeauthor{vaswani2017attention}~\shortcite{vaswani2017attention}, the dropout \cite{srivastava2014dropout} with the keep probability of $p_{ad}$ can be applied to both the token2token and the source2token attention probabilities in MTSA. In particular, the $\dropout$ can be applied to each of two matrices composing the dividend in Algorithm 1 Step 7, i.e., replacing ``$E_c^X E_c^R$'' with ``$\dropout (E_c^X) \dropout(E_c^R)$'', each with the keep probability of $\sqrt{p_{ad}}$. 

\section{Training Setups} \label{app:training}

The optimization objectives for classification and regression problems are cross-entropy loss and mean square error respectively, which we minimize by using Adadelta \cite{zeiler2012adadelta} or Adam \cite{kingma2014adam} optimizer. All trainable weight matrices are initialized by Glorot Initializer \cite{glorot2010understanding}, and all the biases are initialized as zeros. We use 300D (except 200D for SRL task) GloVe 6B pre-trained vectors \cite{pennington2014glove} to initialize the word embeddings. The embedding for a word in the training set but not in GloVe is randomly initialized by sampling from uniform distribution between $[-0.05, 0.05]$. The word embeddings will be fine-tuned during the training phase. We also apply Dropout with keep probability $p_{kp}$, and L2 regularization with weight decay factor $\gamma$ to all the model for avoiding overfitting. The unspecified activation functions for all fully-connected layers appearing in models are set to $\relu$ \cite{glorot2011deep}. The activation function $\sigma _t (\cdot)$ applied to token2token alignment scores is set to $\log (\sigmoid(\cdot))$ unless otherwise specified. 

\textbf{For fair and reliable comparisons with baselines and prior works}, on SNLI and sentence classification tasks, we follow training setup and hyperparameters used in corresponding prior works, and only tune the dropout probability for different baseline or ablation models, without any other trick (e.g., learning rate schedule, batch/layer normalization, etc.); on SRL, we directly employ the training setup and the well-tuned hyperparameters used in the prior state-of-the-art work \cite{tan2017deep} based on multi-head self-attention mechanism, without tuning them specifically for our proposed model. Besides, for the language model based transfer learning for SNLI and MultiNLI tasks, we use the pretrained model provided by \citeauthor{radford2018improving}~\shortcite{radford2018improving}. And, we use the language model as the auxiliary task for models' universality with the coefficient of $0.3$, and use other hyper-parameters (e.g., initial learning rate, optimizer, leaning rate schedule, epoch number) given by \citeauthor{radford2018improving}~\shortcite{radford2018improving}. 

Finally, We give the details about hyper-parameter selection which leads the proposed model to achieve the optimal performance for each NLP benchmark in the following.

For SNLI dataset (natural language inference), we set $p_{kp} = 0.65$ and $\gamma = 5\times10^{-5}$, and use Adadelta as the optimizer with mini batch size of $128$. And, we do not use the attention dropout for this benchmark. Besides, the activation function for fully-connected layer is set to $\elu$ \cite{clevert2016fast}.  The training procedure is completed within 600K steps, approximately costing 12 hours.

For CoNLL-05 dataset (semantic role labeling), we use the same hyper-parameters provided in \cite{tan2017deep} rather than tune them for a fair comparison. The keep probabilities of dropout for fully-connected layer and residual connection~\cite{he2016deep} are set to $0.9$ and $0.8$ respectively. The attention dropout with keep probability of $0.9$ is applied to both source2token and token2token alignment scores, which equals to setting the probability to $0.81$ in MTSA. And, the activation function $\sigma_t (\cdot)$ applied to the token2token alignment scores is set to $\identity(\cdot)$. Besides, different from using fixed positional embedding in \cite{tan2017deep}, we remove it and only use the forward and backward masks in MTSA to encode bi-directional order information. The training will finish within about 20 hours by using Adadelta optimizer.

For CR, MPQA and SUBJ datasets, we set $p_{kp} = 0.5$ and $\gamma = 10^{-4}$ for these three benchmarks. And we apply attention dropout with keep probability of $0.8$ to CR and MPQA. Different from the other experiments in this paper, we here use Adam as the optimizer to train the models, and do not use any learning rate decay trick. The training procedure is completed within 1000 batch steps.

For TREC dataset (question-type classification), we set $p_{kp} = 0.5$ and $\gamma = 10^{-4}$ and do not apply the attention dropout. The  training procedure is completed within 80K steps by using Adadelta optimizer.

For SST-5 dataset (sentiment analysis), we set $p_{kp} = 0.7$ and $\gamma = 10^{-4}$ and do not apply the attention dropout. The training procedure is completed within 120K steps by using Adadelta optimizer.

\subsection{Evaluation on Machine Translation} \label{app:training_mt}

For machine translation, due to the limited computation resources, we build two training and decoding setups which require fewer GPUs to fairly and reliably compare the Transformer with either multi-head self-attention or the proposed MTSA. 

According to the reproductivity experiments at \href{https://github.com/tensorflow/tensor2tensor/issues/317#issuecomment-331953713}{issue\#317} in which \emph{transformer\_base} model from official implementation \href{https://github.com/tensorflow/tensor2tensor}{tensor2tensor} needs $8\times\text{P100}$, batch size of 4096 and training step of 250K to achieve the BLEU value of 27.76, our reproductivity experiment of the Transformer with $4\times\text{P100}$, batch size of 6144 and training step of 133K to achieve BLEU value of $\sim$27 is reasonable and accurate. 
The \href{https://github.com/tensorflow/tensor2tensor/issues/444#issuecomment-351391778}{issue\#444} of \href{https://github.com/tensorflow/tensor2tensor}{tensor2tensor} also demonstrates that the Transformer trained on $4\times\text{P100}$ hurts $>\!1$ BLEU point compared to that trained on $8\times\text{P100}$, and the Transformer trained on fewer GPUs cannot achieve state-of-the-art decoding performance even if using more GPU hours. 

\section{Related Work}\label{app:related}

The self-attention mechanism was firstly applied to NLP tasks to implicitly model the syntactic dependency by using a pairwise compatibility function. 
\citeauthor{kim2017structured}~\shortcite{kim2017structured} proposed a syntactic attention mechanism to simulate syntactic tree selection, which can be regarded as a self-attention mechanism making soft-selection based on the learned syntactic dependency model. 
\citeauthor{hu2017reinforced}~\shortcite{hu2017reinforced} presented a self aligning layer to align information between context words, allowing crucial clues to be fused into the context-aware representation, which mitigates a limitation of the capability of a RNN in long-term dependency modeling.
\citeauthor{vaswani2017attention}~\shortcite{vaswani2017attention} proposed a scaled dot-product attention mechanism where a scaled dot-product compatibility function is leveraged to model the syntactic pairwise dependency. They then presented a multi-head attention mechanism based on the dot-product attention, which employs multiple subspaces to capture diverse dependencies and save memory/computation/parameters. An attention-only model, dubbed ``Transformer'', based solely on multi-head attention was finally proposed for sequence to sequence tasks. 
\citeauthor{shen2017disan}~\shortcite{shen2017disan} proposed a multi-dimensional compatibility function to capture feature-level dependency or relevance for attention mechanism. They then introduced a directional (masked) self-attention mechanism, in which the multi-dim compatibility function is used to model the pairwise dependency, and forward and backward positional masks are leveraged to capture bi-directional order information.

Furthermore, there was another type of self-attention mechanism capturing the contribution and dependency of each token to the entire source sequence for a specific task, which can be used on sentence encoding or sequence compression task.
\citeauthor{liu2016learning}~\shortcite{liu2016learning} proposed an intra-sentence attention mechanism where the pooling result of the input sequence is used as the query attending to each token from the same sequence. They applied it to sentence embedding tasks. 
\citeauthor{lin2017structured}~\shortcite{lin2017structured} proposed a self-attentive model using a matrix to represent the sentence embedding, with each row of the matrix attending to a different part of the sentence. It shares a similar idea with the multi-head attention \cite{vaswani2017attention}. 
\citeauthor{shen2017disan}~\shortcite{shen2017disan} proposed a source2token self-attention mechanism that removes the query from the multi-dim compatibility function, for the purpose of directly modeling the feature-wise contribution of each token to the entire input source on a specific task.

Self-attention mechanisms introduced above have been implemented on a wide range of practical tasks and achieved state-of-the-art performance.  
\citeauthor{lin2017structured}~\shortcite{lin2017structured} applied the self-attention model in conjunction with Bi-LSTM to sentence embedding tasks. 
\citeauthor{hu2017reinforced}~\shortcite{hu2017reinforced} integrated the self aligning layer with general query-context mutual-attention framework (i.e., BiDAF \cite{seo2017bidirectional}) to model long-term dependency for machine comprehension task.  
\citeauthor{vaswani2017attention}~\shortcite{vaswani2017attention} applied the attention-only sequence to sequence model, ``Transformer'', to neural machine translation. 
\citeauthor{shen2017disan}~\shortcite{shen2017disan} employed the directional and source2token self-attention mechanisms respectively as context fusion and sequence compression modules to build a sentence embedding model.  
\citeauthor{tan2017deep}~\shortcite{tan2017deep} applied the stacked multi-head self-attention mechanism jointly with fully-connected layer (similar to the encoder in Transformer) to the semantic role labeling task. 
\citeauthor{im2017distance}~\shortcite{im2017distance} proposed distance-aware masks (sharing a similar idea with directional self-attention) to model the distance information between every two tokens in a sequence, and applied it to sentence-encoding based natural language inference task. 
\citeauthor{liu2018generating}~\shortcite{liu2018generating} facilitated the passage summarization problem to a language model problem, and used the decoder from Transformer to solve this problem. 
\citeauthor{yu2018qanet}~\shortcite{yu2018qanet} employed stacked CNN and self-attention mechanism to model local and long-term dependencies respectively, and achieved new state-of-the-art performance on machine comprehension task. 
\citeauthor{velivckovic2017graph}~\shortcite{velivckovic2017graph} implemented a stacked multi-head attention on a graph to perform transductive and inductive graph tasks, where a node is used as the query attending to its neighboring nodes.

\begin{figure*} [htbp]
	\vspace{-5mm}
	\centering
	\begin{minipage}[c]{.23\textwidth}
		\centering
		\includegraphics[width=\textwidth]{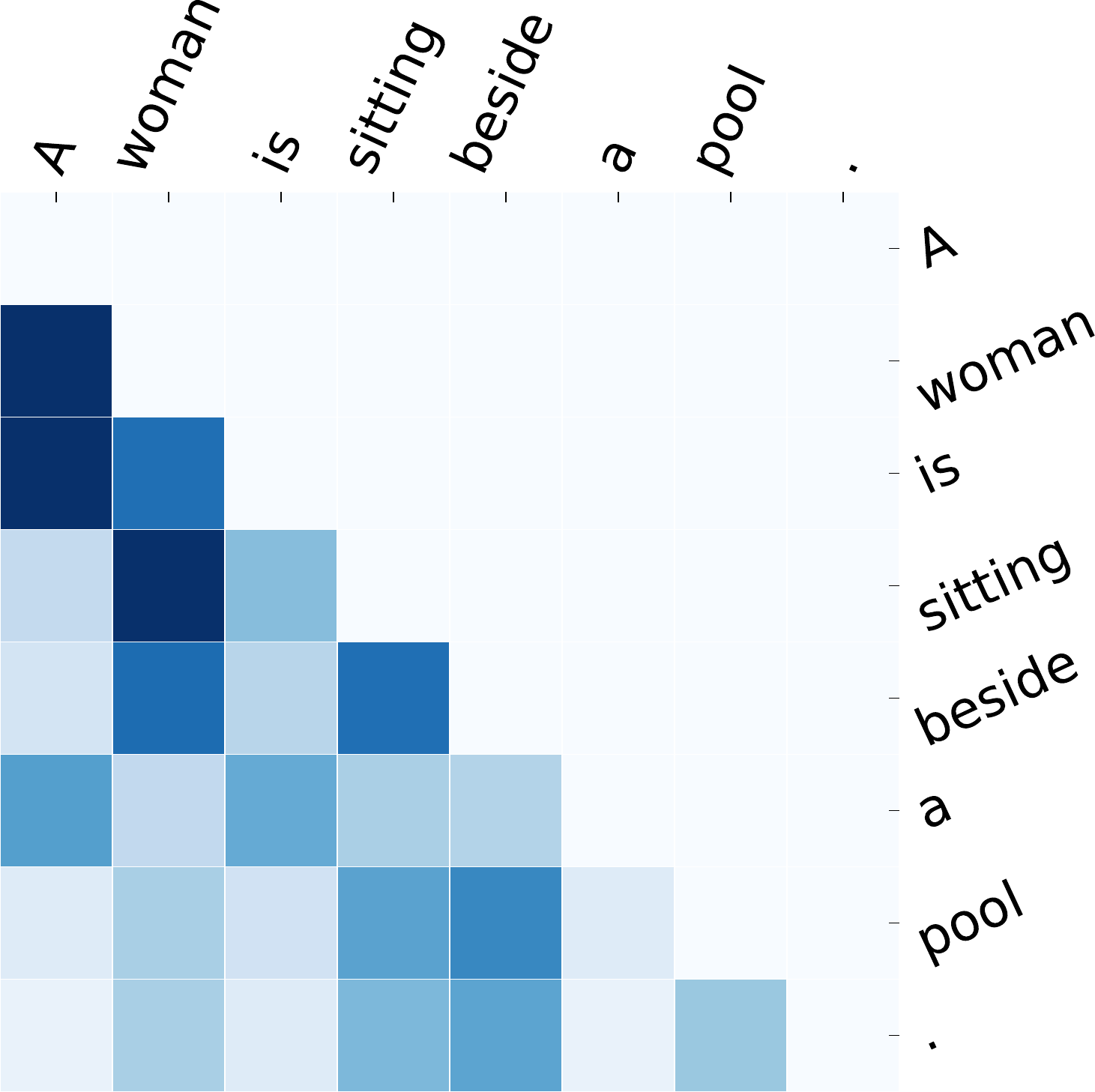}
	\end{minipage}
	\begin{minipage}[c]{.23\textwidth}
		\centering
		\includegraphics[width=\textwidth]{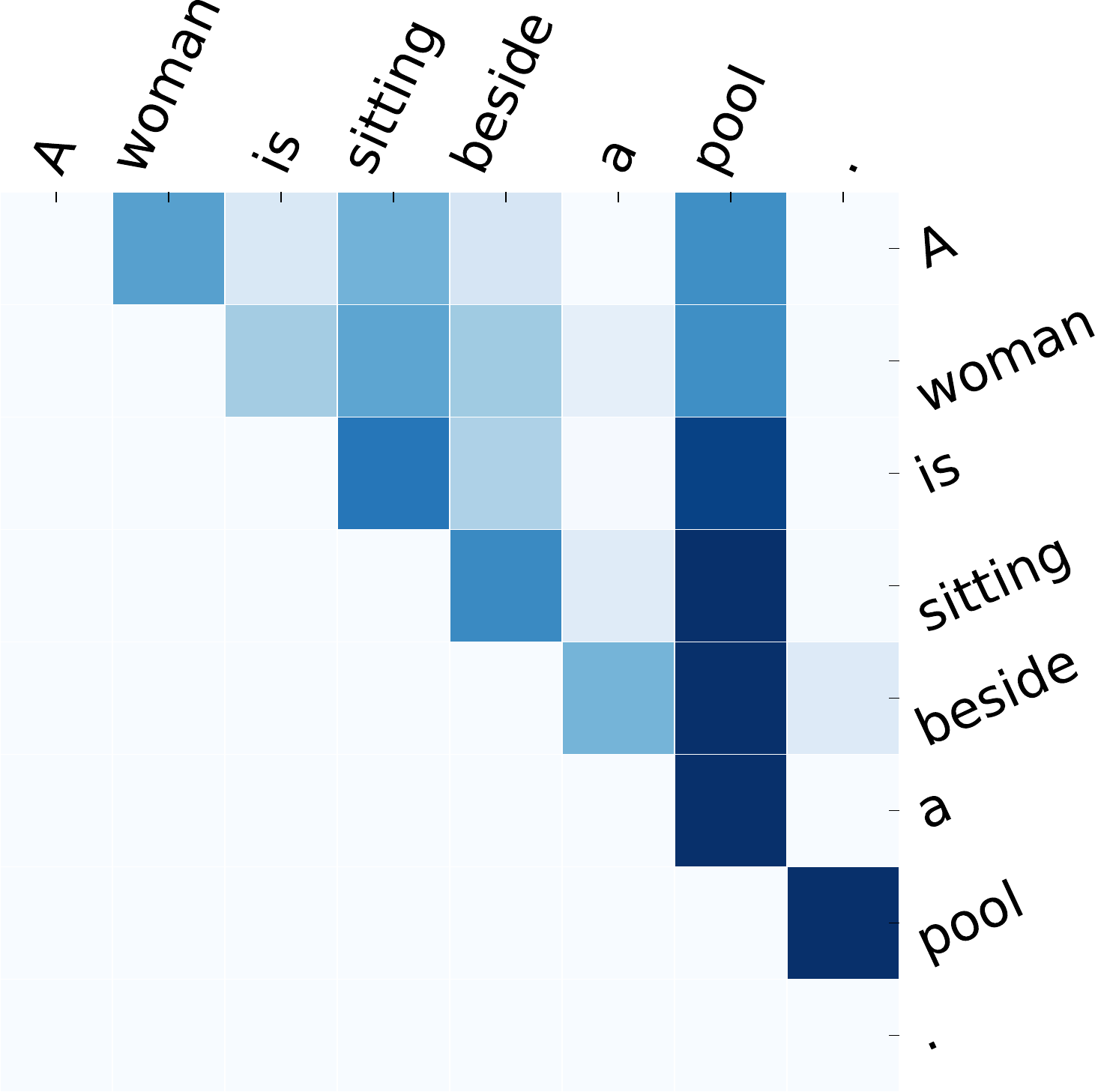}
	\end{minipage}
	\begin{minipage}[c]{.24\textwidth}
		\centering
		\includegraphics[width=\textwidth]{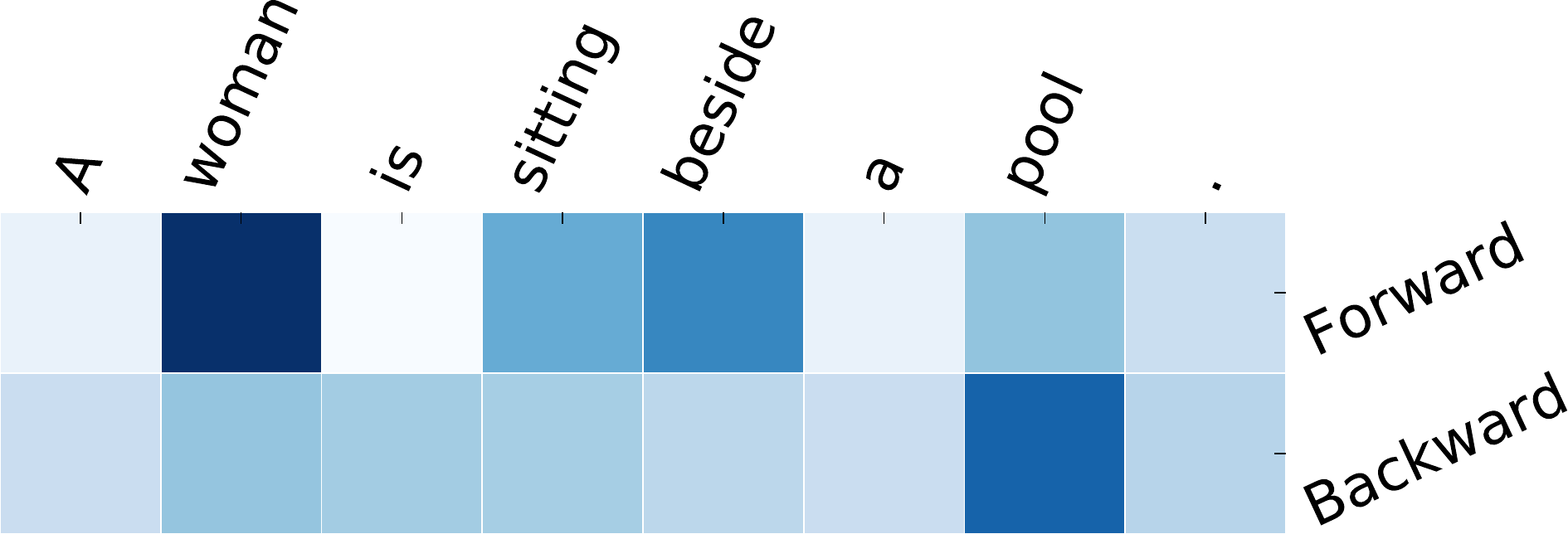}
	\end{minipage} 
	\vspace{0mm} \\
	\begin{minipage}[c]{.24\textwidth}
		\centering
		\includegraphics[width=\textwidth]{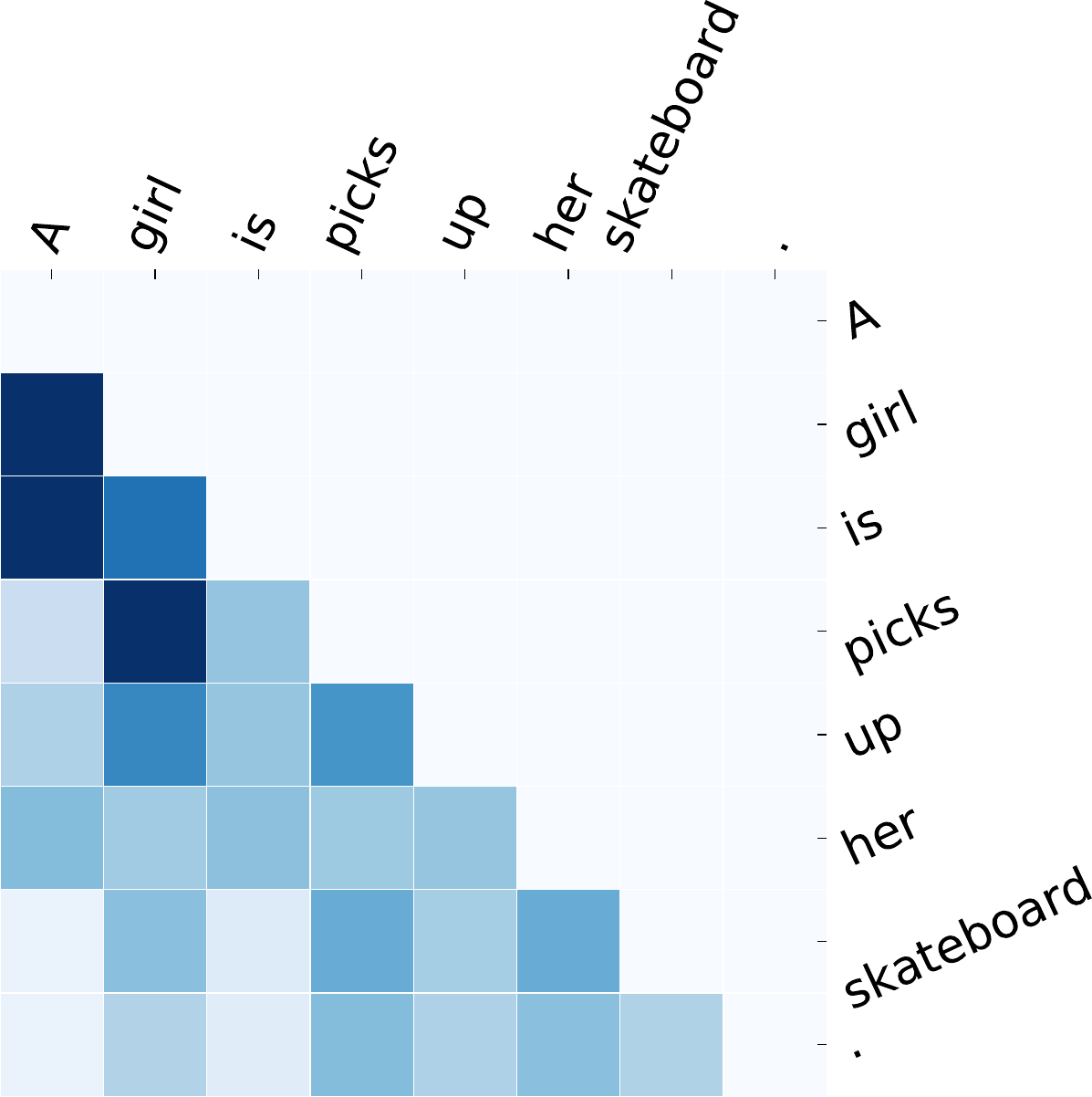}
	\end{minipage}
	\begin{minipage}[c]{.24\textwidth}
		\centering
		\includegraphics[width=\textwidth]{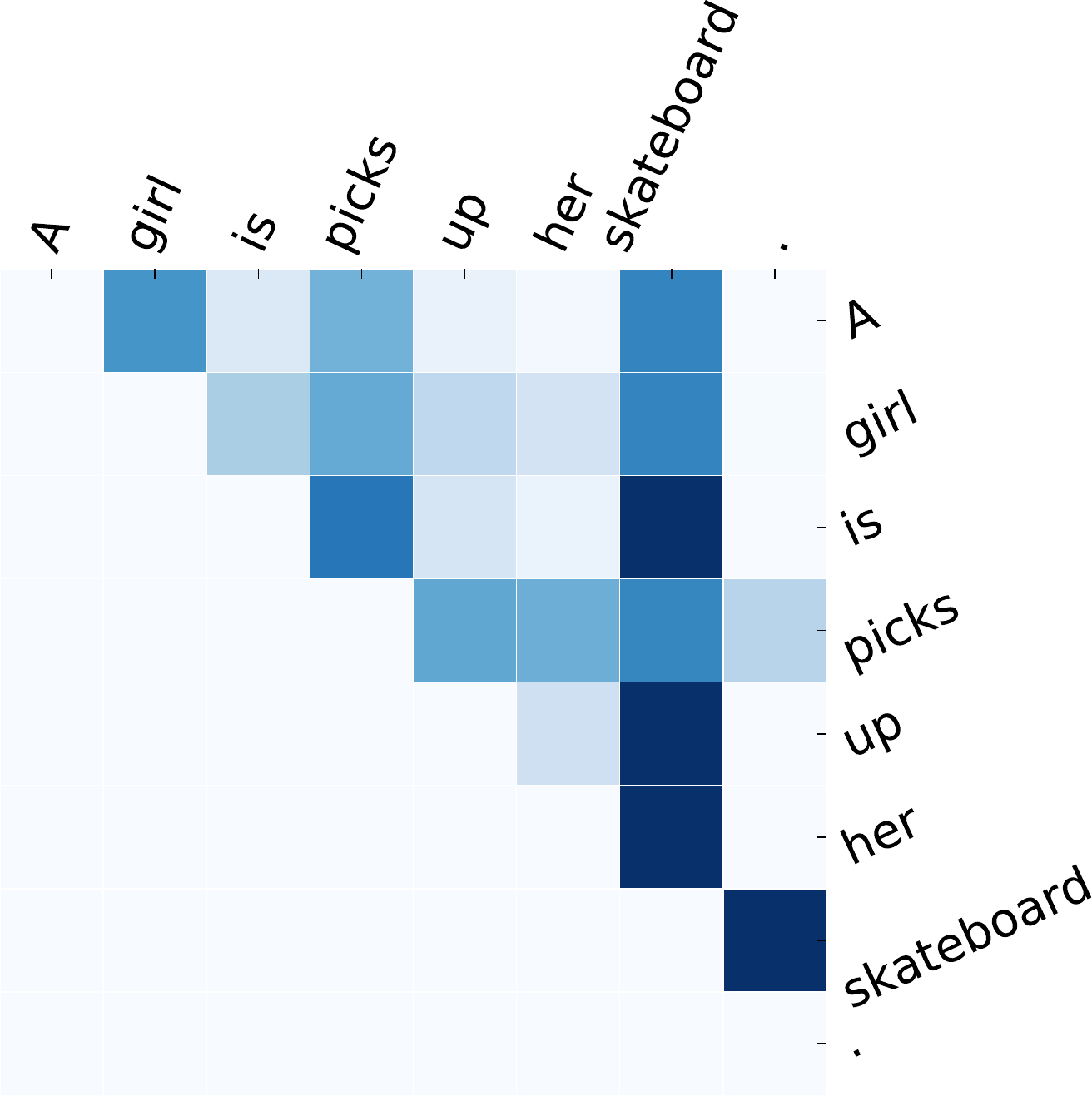}
	\end{minipage}
	\begin{minipage}[c]{.26\textwidth}
		\centering
		\includegraphics[width=\textwidth]{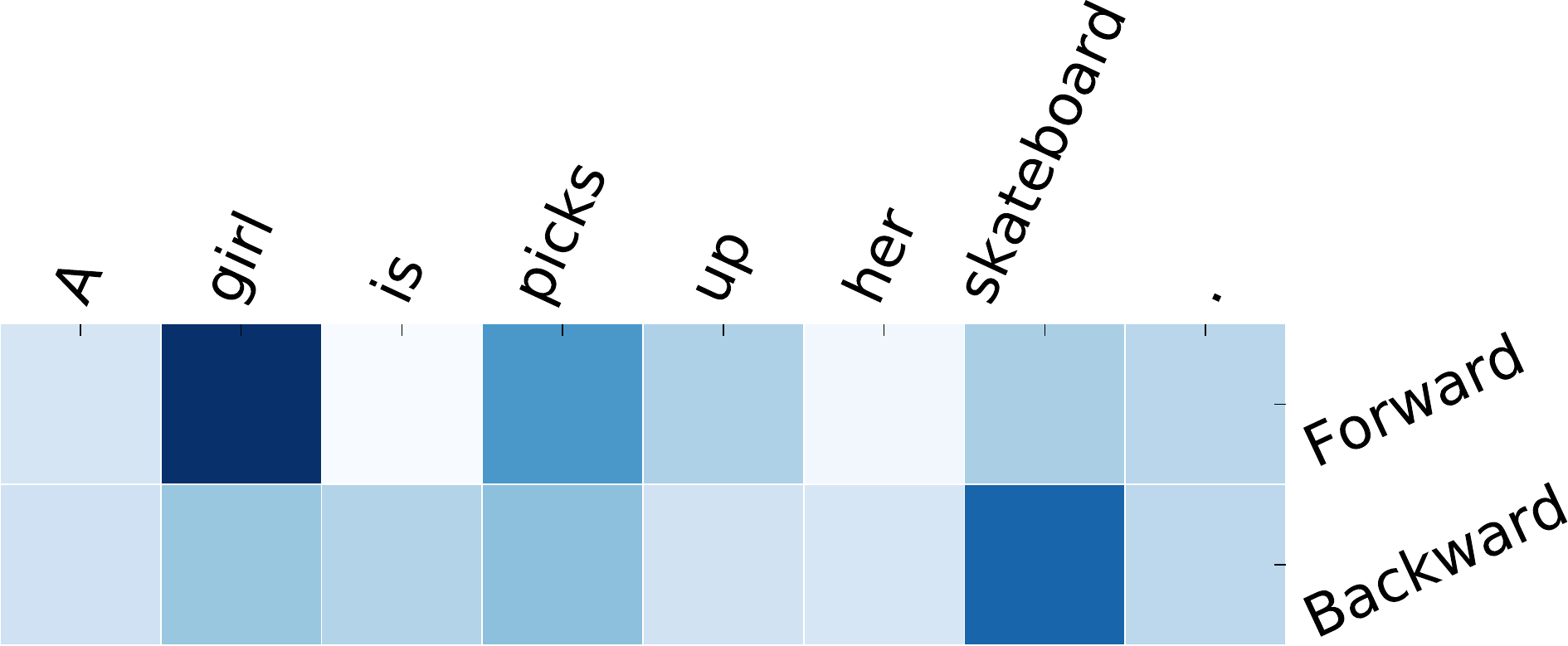}
	\end{minipage} 
	\vspace{0mm} \\
	\begin{minipage}[c]{.28\textwidth}
		\centering
		\includegraphics[width=\textwidth]{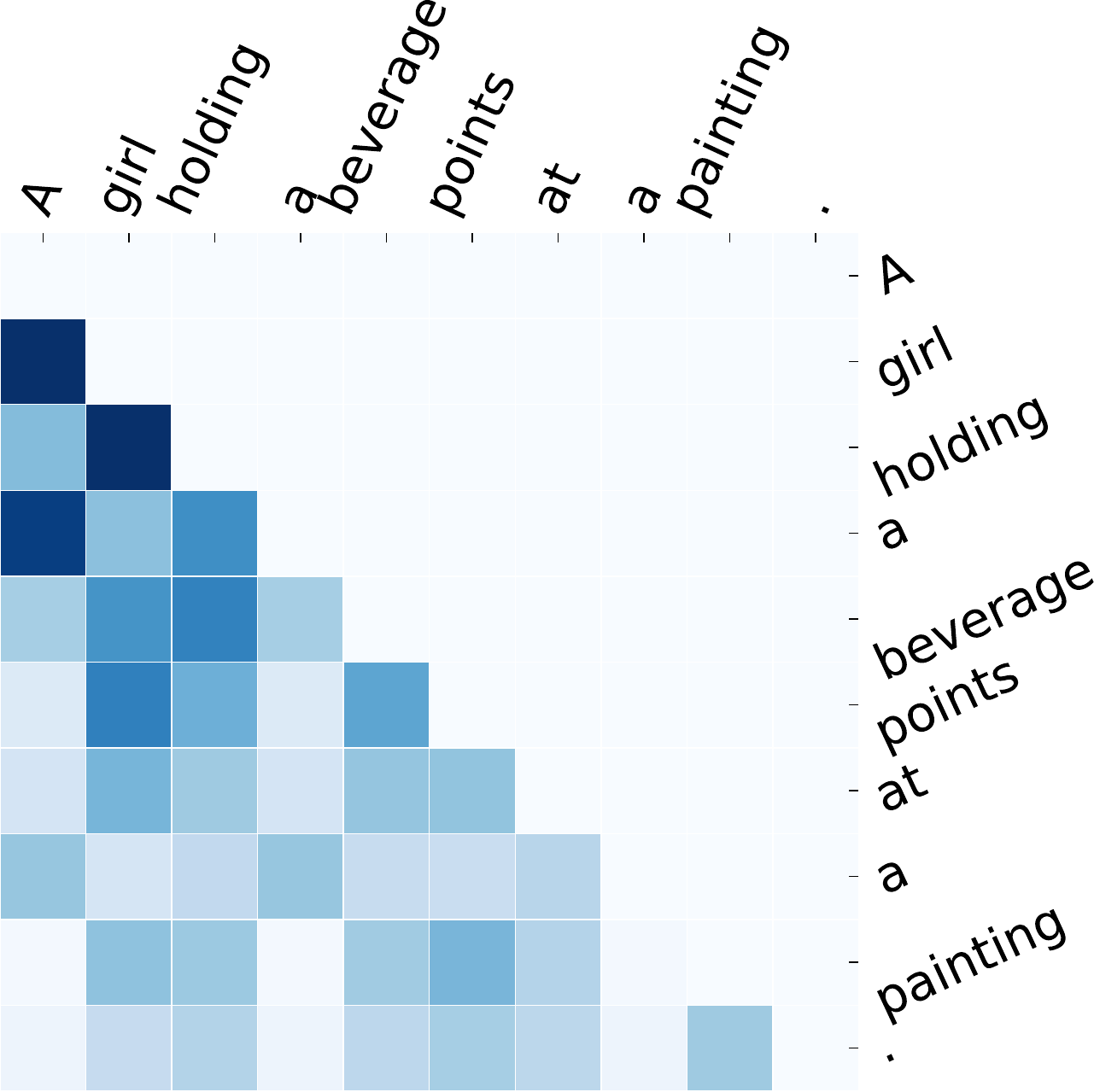}
	\end{minipage}
	\begin{minipage}[c]{.28\textwidth}
		\centering
		\includegraphics[width=\textwidth]{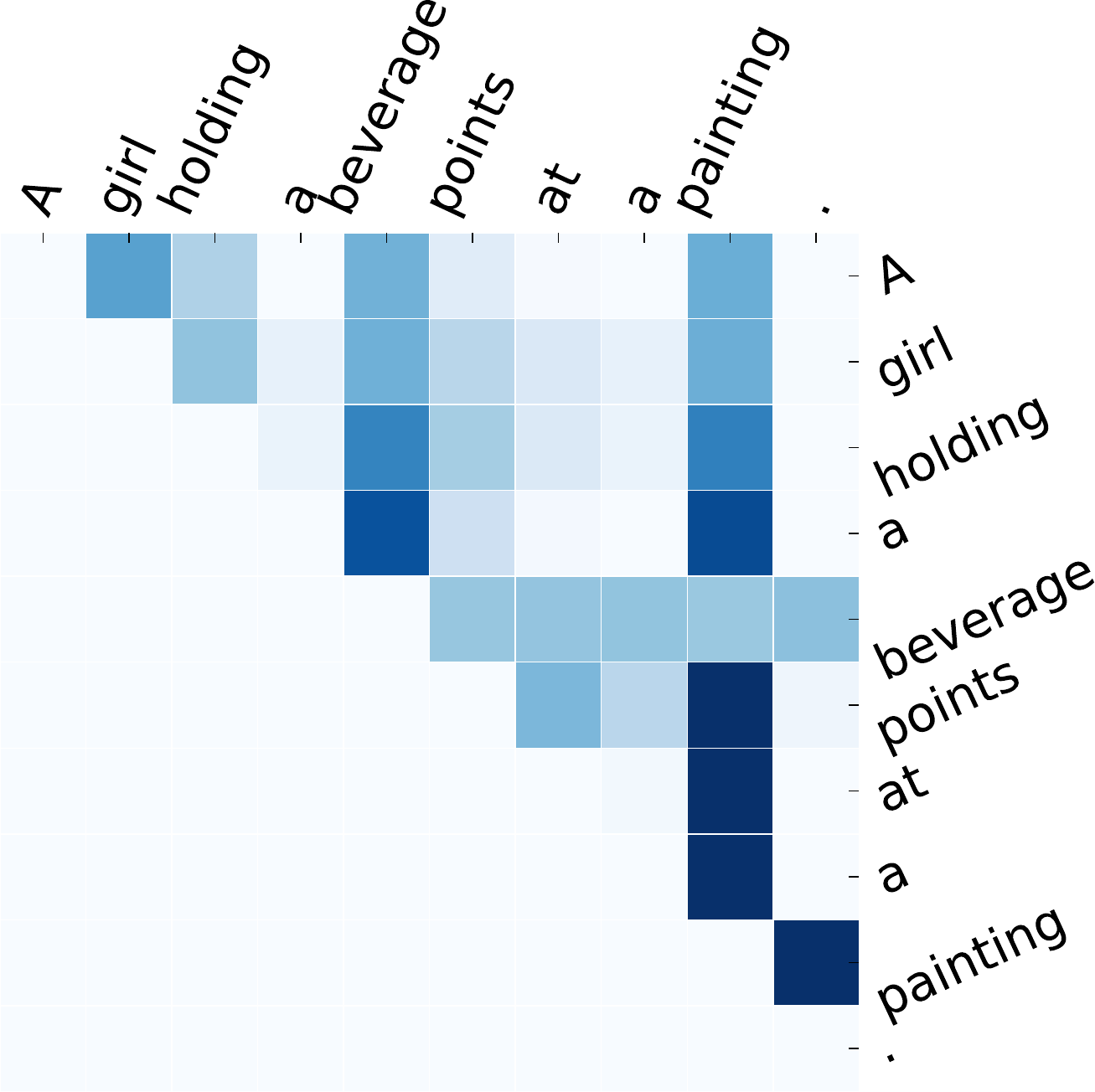}
	\end{minipage}
	\begin{minipage}[c]{.3\textwidth}
		\centering
		\includegraphics[width=\textwidth]{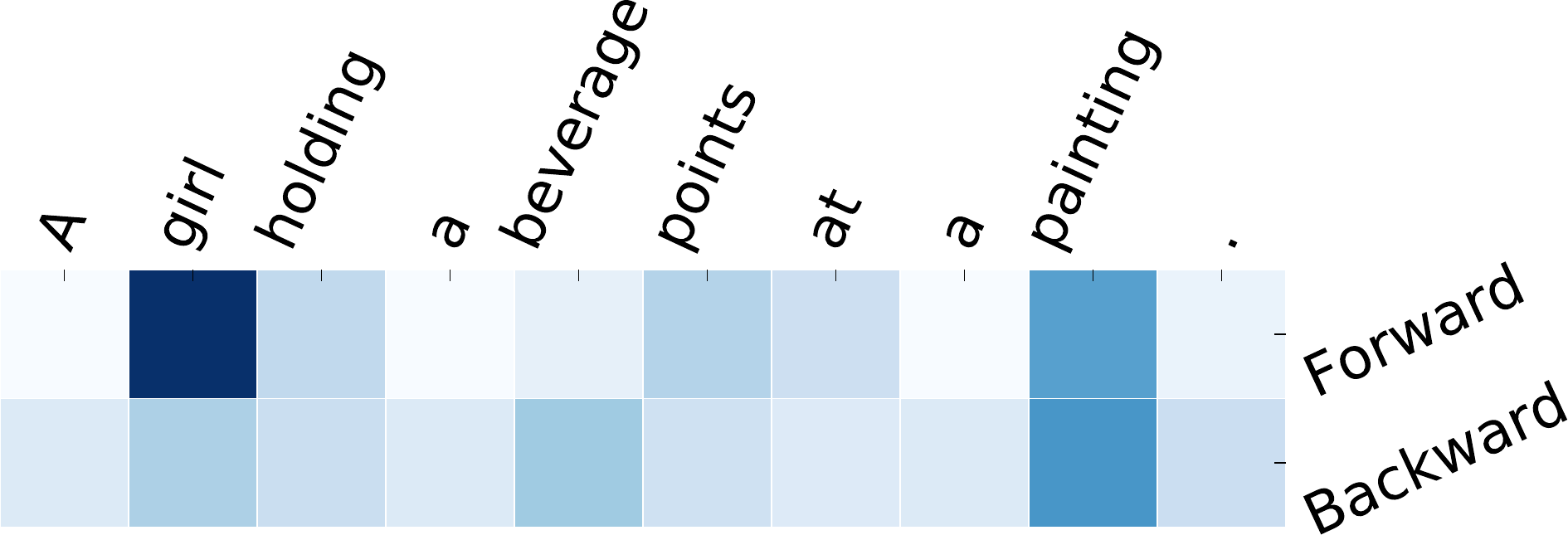}
	\end{minipage} 
	\vspace{0mm} \\
	\begin{minipage}[c]{.28\textwidth}
		\centering
		\includegraphics[width=\textwidth]{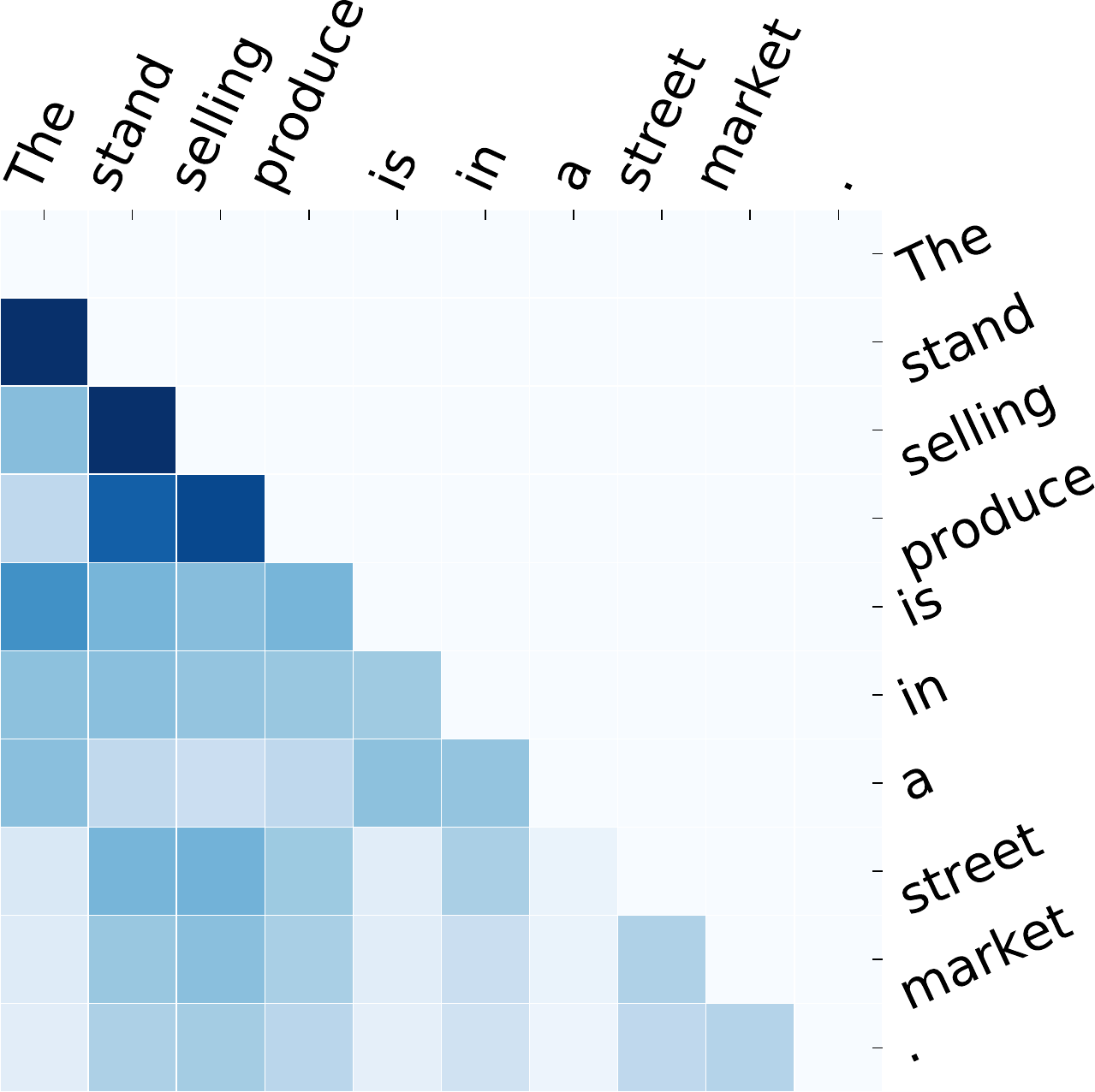}
	\end{minipage}
	\begin{minipage}[c]{.28\textwidth}
		\centering
		\includegraphics[width=\textwidth]{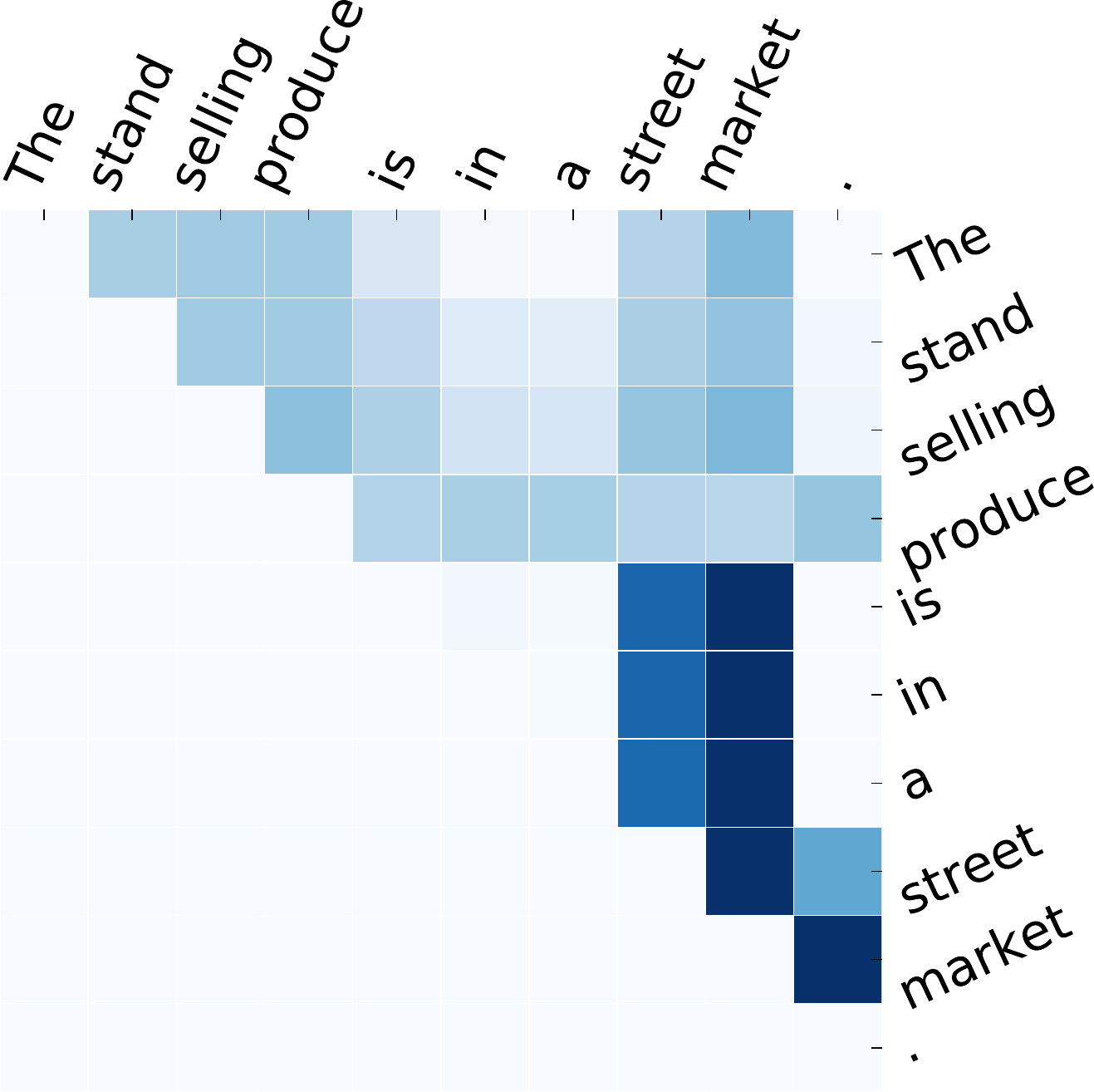}
	\end{minipage}
	\begin{minipage}[c]{.3\textwidth}
		\centering
		\includegraphics[width=\textwidth]{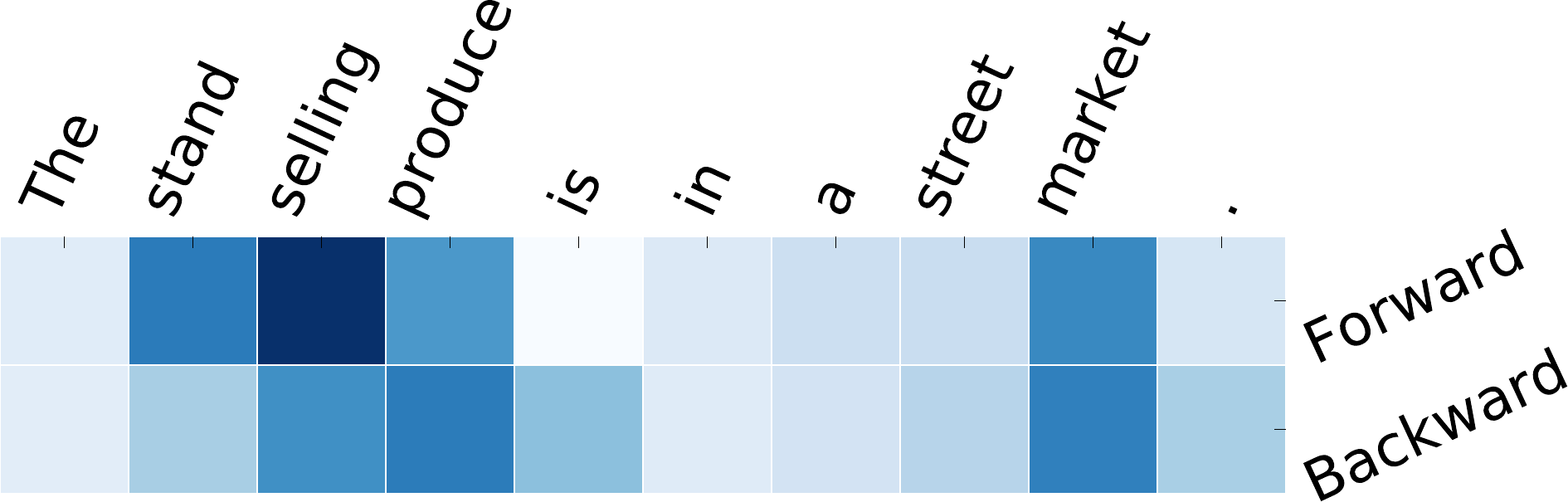}
	\end{minipage} 
	\vspace{0mm} \\
	\begin{minipage}[c]{.28\textwidth}
		\centering
		\includegraphics[width=\textwidth]{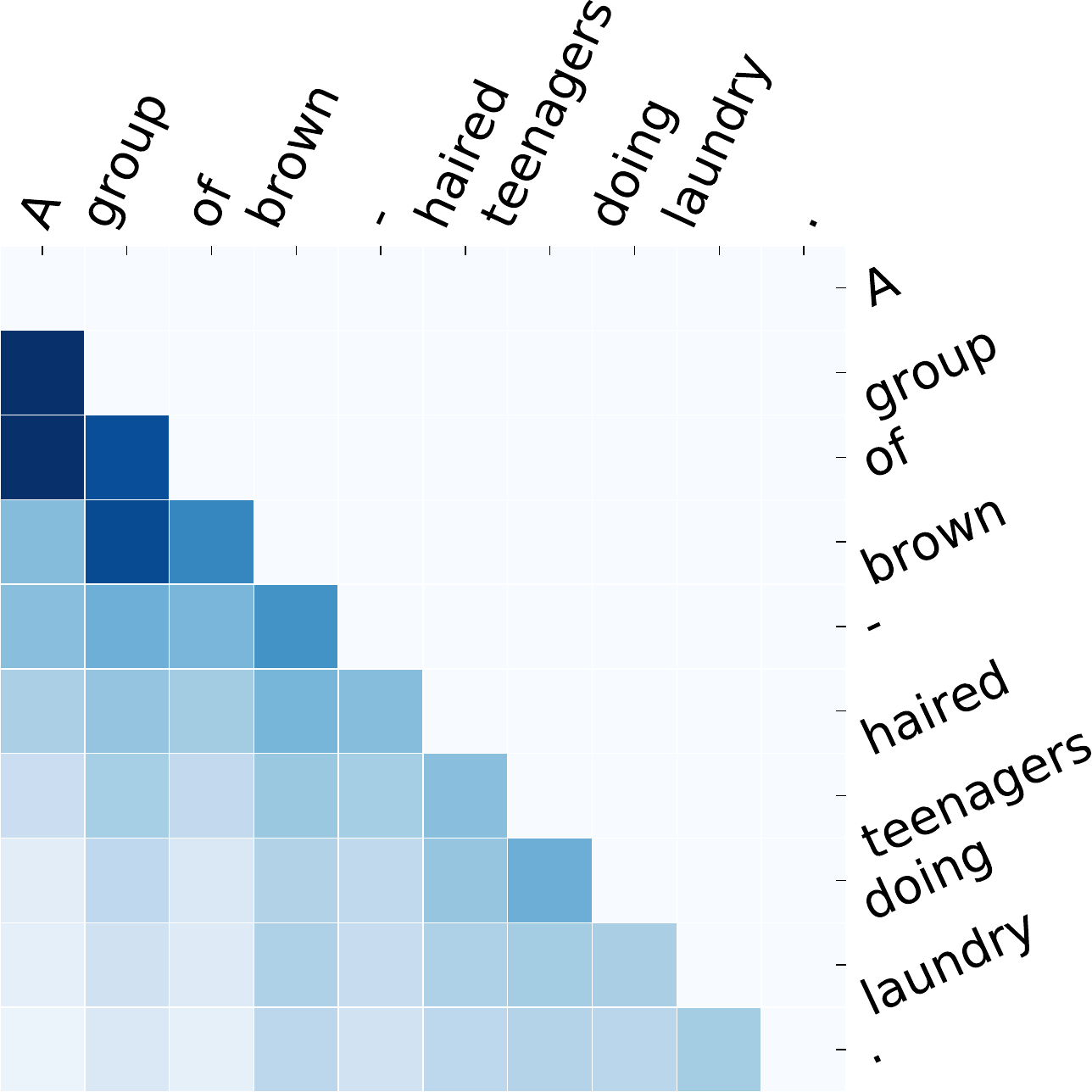}
	\end{minipage}
	\begin{minipage}[c]{.28\textwidth}
		\centering
		\includegraphics[width=\textwidth]{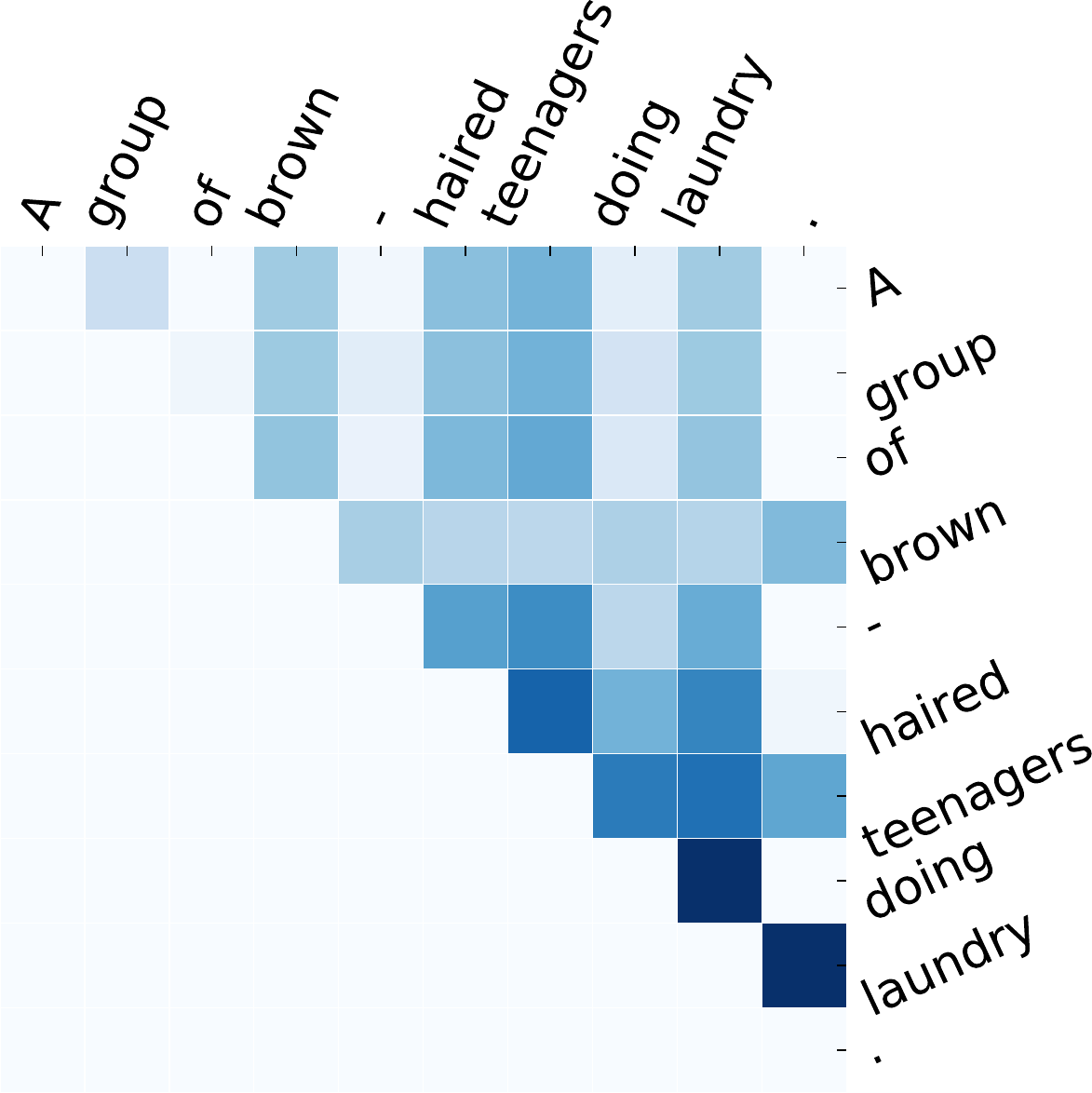}
	\end{minipage}
	\begin{minipage}[c]{.3\textwidth}
		\centering
		\includegraphics[width=\textwidth]{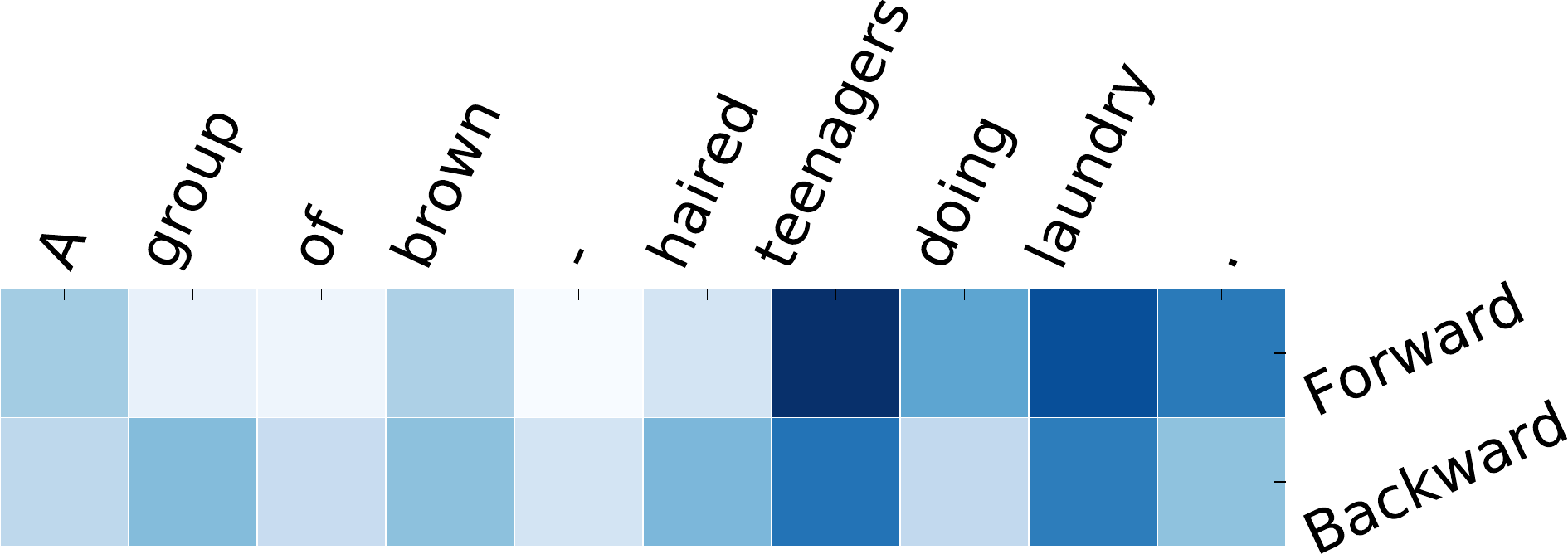}
	\end{minipage} 
	\caption{\small Heatmaps for normalized token2token and source2token alignment scores with forward and backward masks. Each row shows three types of scores associated with the tokens from a same sentence: token2token alignment scores in TSA with forward mask (left), token2token alignment scores in TSA with backward masks (middle), source2token alignment scores at token level for the two heads with forward and backward masks (right). The tokens in $x$-axis and $y$-axis are the dependent and governor tokens, respectively.}
	\label{fig:vis_attn}
\end{figure*}

\section{Discussion} \label{app:discussion}

In this paper, we propose a novel multi-dim self-attention mechanism, called multi-mask tensorized self-attention (MTSA), for context fusion purpose. It is equipped with an expressive but previously inefficient multi-dim compatibility function to compute tensorized alignment scores that can capture both pairwise and global dependencies. However, it does not suffer from any time or memory explosion problem that precludes previous multi-dim attention mechanisms from being applied to large-scale datasets or long sequences. Meanwhile, multiple distinct positional masks are applied to multiple heads (subspaces) to model different types of sequential and structural information of input sequence. 
The experimental results show that MTSA empirically achieves state-of-the-art performance on a wide range of NLP tasks, and is as fast and as memory-efficient as CNN baselines. This indicates that a stacked version of MTSA might improve the performance on more NLP tasks.

There are various intriguing works that are worth exploring based on the proposed model, such as 1) integrating MTSA with Transformer \cite{vaswani2017attention} for more complicated and high-level NLP tasks (e.g., neural machine translation and summarization), 2) applying more types of positional masks or distance-aware masks \cite{im2017distance} to different heads, and thus taking into account richer structure information, and 3) integrating the light-weight and time-efficient MTSA with a hierarchical self-attention structure \cite{shen2018biblosan} for context fusion on long text (e.g., passage and document), rather than using single self-attention mechanism or the non-parallelizable RNN-based models.

\section{Visualization} \label{app:vis}

In this section, we use heatmaps to visualize the token2token and source2token alignment scores computed by MTSA mechanism with forward and backward positional masks. The sentences used for visualization are randomly selected from the test set of SNLI dataset. 
For clarity, the visualized token-level alignment score of multi-dim source2token self-attention is computed by averaging the corresponding vector of feature-wise alignment scores.

As shown in Figure \ref{fig:vis_attn}, the heatmaps of the alignment scores computed by MTSA mechanism demonstrate that,
1) a token pair with strong syntactic relevance is assigned with high alignment score by the pairwise compatibility function;
2) the important tokens (e.g., verbs and nouns ) usually achieve high source2token alignment scores, whereas the trivial tokens (e.g., stop words) obtain relatively low alignment scores; and 
3) MTSA mechanism with backward and forward masks focuses on different positions of the input sentence in different heads (subspaces), which makes the final attention results more versatile and diverse.

\end{document}